\documentclass{article}

\usepackage[main, final]{neurips_2026}

\usepackage[utf8]{inputenc}
\usepackage[T1]{fontenc}
\usepackage{amsmath}
\usepackage{amssymb}
\usepackage{amsfonts}
\usepackage{bm}
\usepackage{nicefrac}

\usepackage{graphicx}
\usepackage{booktabs}
\usepackage{multirow}
\usepackage{colortbl}
\usepackage[flushleft]{threeparttable}
\usepackage{subcaption}
\usepackage{caption}
\usepackage{wrapfig}

\usepackage{booktabs}
\usepackage{multirow}
\usepackage{makecell}
\usepackage{colortbl}
\usepackage[flushleft]{threeparttable}
\usepackage{amssymb}

\usepackage{xspace}
\usepackage{microtype}
\usepackage{soul}
\usepackage{pifont}
\usepackage{esvect}
\usepackage{enumitem}
\usepackage{setspace}

\usepackage{xcolor}
\definecolor{ceruleanblue}{rgb}{0.16, 0.32, 0.75}

\usepackage[
    pagebackref,
    breaklinks,
    colorlinks,
    citecolor=ceruleanblue,
    linkcolor=ceruleanblue,
    urlcolor=ceruleanblue
]{hyperref}

\usepackage[capitalize]{cleveref}

\usepackage[toc,page]{appendix}
\usepackage{minitoc}

\crefname{section}{Sec.}{Secs.}
\Crefname{section}{Section}{Sections}

\crefname{table}{Tab.}{Tabs.}
\Crefname{table}{Table}{Tables}

\crefname{figure}{Fig.}{Figs.}
\Crefname{figure}{Figure}{Figures}

\crefname{appendix}{Appendix}{Appendices}
\Crefname{appendix}{Appendix}{Appendices}
\usepackage{multirow}

\definecolor{lightgray}{RGB}{240,247,255}
\definecolor{car}{RGB}{230,25,75}
\definecolor{cyclist}{RGB}{255, 130, 48}
\definecolor{pedestrian}{RGB}{138, 43, 226}
\definecolor{colorgt}{RGB}{170,255,195}
\definecolor{white}{rgb}{1.0, 1.0, 1.0}
\definecolor{monocop}{RGB}{71, 159, 179}
\definecolor{baseline}{RGB}{253, 190, 110}
\definecolor{gray}{RGB}{243, 243, 243}
\definecolor{darkGreen}{rgb}{0.01, 0.8, 0.24}
\definecolor{hdColor}{rgb}{0.9,0.95,1}

\definecolor{myblue}{HTML}{1C5CB8}
\definecolor{myred}{RGB}{180,0,0}

\newcommand{\monoThreeD}{Mono3D\xspace}

\newcommand{\twoD}{$2$D\xspace}
\newcommand{\threeD}{$3$D\xspace}

\newcommand{\iouThreeD}{IoU$_{3\text{D}}$\xspace}

\newcommand{\bev}{BEV\xspace}

\newcommand{\mathDash}{$-$}

\newcommand{\kitti}{KITTI\xspace}
\newcommand{\nuscenes}{nuScenes\xspace}
\newcommand{\waymo}{Waymo\xspace}

\newcommand{\omniThreeD}{Omni3D\xspace}

\def\omnidataset{Omni3D\xspace}

\def\omnidatasetout{Omni3D$_{\text{OUT}}$\xspace}
\def\omnidatasetkitti{Omni3D$_{\text{KITTI}}$\xspace}
\def\omnidatasetnu{Omni3D$_{\text{nuScenes}}$\xspace}
\def\omnidatasetoutlarge{Omni3D$_{\text{OUT}}^{\text{Large}}$\xspace}
\def\omnidatasetoutsmall{Omni3D$_{\text{OUT}}^{\text{Small}}$\xspace}
\def\omnidatasetoutcar{Omni3D$_{\text{OUT}}^{\text{Car}}$\xspace}

\newcommand{\valOne}{Val\xspace}

\newcommand{\val}{Val\xspace}

\newcommand{\test}{Test\xspace}

\newcommand{\ap}{AP}

\newcommand{\apThreeD}{\ap$_{3\text{D}}$\xspace}
\newcommand{\aphThreeD}{APH$_{3\text{D}}$\xspace}

\newcommand{\apBev}{\ap$_{\text{BEV}}$\xspace}

\newcommand{\sota}{SoTA\xspace}

\newcommand{\deviant}{DEVIANT\xspace}
\newcommand{\monoflex}{MonoFlex\xspace}

\newcommand{\gupNet}{GUP Net\xspace}

\newcommand{\monodetr}{MonoDETR\xspace}
\newcommand{\monodgp}{MonoDGP\xspace}

\newcommand{\occupancymThreeD}{OccupancyM3D\xspace}
\newcommand{\opaThreeD}{OPA-3D\xspace}

\newcommand{\monocon}{MonoCon\xspace}
\newcommand{\fdThreeD}{FD3D\xspace}
\newcommand{\monorcnn}{MonoRCNN\xspace}
\newcommand{\monouni}{MonoUNI\xspace}
\newcommand{\monocd}{MonoCD\xspace}
\newcommand{\monomae}{MonoMAE\xspace}
\newcommand{\monocop}{MonoCoP\xspace}

\newcommand{\monotakd}{MonoTAKD\xspace}

\newcommand{\eg}{\textit{e.g.}}

\providecommand{\RHD}{\blacktriangleright}
\providecommand\rightarrowRHD{\relbar\joinrel\mathrel\RHD}

\newcommand{\uparrowRHD}  {\rotatebox[origin=c]{90}{$\rightarrowRHD$}}

\newcommand{\uparrowRHDSmall}  {\raisebox{0.05\normalbaselineskip}{\scalebox{0.7}{\uparrowRHD}}}

\usepackage{pifont}
\newcommand{\noIndentHeading}[1]{\noindent\textbf{#1}}

\definecolor{XLcolor}{rgb}{0.858, 0.188, 0.478}

\title{Rethinking Geometric Depth in Monocular 3D Object Detection: A Projection-Consistent Reformulation}

\author{%
  Zhihao Zhang$^{1}$ \quad Abhinav Kumar$^{1}$ \quad Huaizhi Qu$^{2}$ \quad Tianlong Chen$^{2}$ \quad Xiaoming Liu$^{1,2}$ \\
  $^{1}$Michigan State University \quad $^{2}$University of North Carolina at Chapel Hill \\
}

\newcommand{\codeURL}{\url{https://github.com/alanzhangcs/MonoPCA}}

\newcommand{\methodName}{MonoPCA\xspace}
\newcommand{\newDepth}{PCDepth\xspace}
\newcommand{\geoDepth}{GeoDepth\xspace}

\newcommand{\mixtureBackbone}{MoB\xspace}
\newcommand{\geoDistill}{GeoAlign\xspace}
\newcommand{\featureAlignment}{HFA\xspace}
\newcommand{\scaleMatch}{SM\xspace}
\newcommand{\scaleAttention}{SA\xspace}

\begin{document}

\maketitle

\begin{abstract}
Monocular \threeD detection (\monoThreeD) hinges on recovering depth from a single image, for which the geometric formulation $z \approx f \cdot H / h_{2D}$ is widely adopted due to its simplicity and direct grounding in projective geometry.
We show that this formulation harbors a systematic, projection-induced bias: substituting the \twoD bounding-box height $h_{2D}$ recovers the depth of an object's visible front surface rather than its center, causing size-dependent underestimation that persists even under ground-truth inputs.
We address this front-surface bias with \methodName, a unified framework built on a projection-consistent reformulation.
At its core, \newDepth retains the measurable $h_{2D}$ and closes the gap to center depth through an orientation- and size-aware correction derived from perspective geometry, eliminating the bias without introducing ill-posed regression targets.
Because \newDepth couples depth to predicted object size and orientation,  attribute perception becomes essential.
We therefore introduce \geoDistill, a teacher--student framework that transfers geometric representations from Vision Foundation Models into a lightweight student, strengthening attribute perception at no inference cost.
Across four challenging benchmarks, \methodName achieves \sota performance, including a $+1.44\%$ \apThreeD gain on the \kitti leaderboard and $+2.80\%$ on \omnidatasetout. Code is available at \mbox{\codeURL}.
\end{abstract}

\section{Introduction}
\label{sec:intro}
Monocular \threeD object detection (\monoThreeD) seeks to recover object location, size, and orientation from a single RGB image, underpinning applications from autonomous driving~\cite{simonelli2020disentangling, ma20233d, ma2022vision} to augmented reality~\cite{merrill2022symmetry, park2021pseudo}.
Yet a single image fundamentally lacks metric scale: projecting a \threeD scene onto a \twoD plane collapses absolute distance into a one-to-many ambiguity.
Depth estimation thus emerges as an ill-posed inverse problem and, in practice, the main bottleneck of \monoThreeD~\cite{ma20233d}.

A substantial body of work has sought to tame this ambiguity, broadly along two complementary axes.
The first injects geometric and probabilistic priors into depth prediction, including keypoint-based reasoning in \monoflex~\cite{zhang2021objects}, geometric decomposition in \monorcnn~\cite{shi2021geometry}, and uncertainty modeling in \gupNet~\cite{lu2021geometry}.
The second reshapes the depth representation itself, through scale-equivariant features in DEVIANT~\cite{kumar2022deviant}, depth-guided decoding in \monodetr~\cite{zhang2023monodetr}, complementary depth heads in \monocd~\cite{yan2024monocd}, decoupled object queries in \monodgp~\cite{pu2024monodgp}, and inter-attribute coupling in \monocop~\cite{zhang2025unleashing}.
Despite their architectural diversity, these methods rest on a shared foundation: when grounded geometrically, depth is recovered through the projection relation $z \approx f \cdot H/h$, with $h$ instantiated as the predicted \twoD bounding-box height $h_{2D}$~\cite{shi2021geometry, lu2021geometry, kumar2022deviant, zhang2023monodetr, pu2024monodgp, zhang2025unleashing}.
This shared foundation, we find, harbors a systematic bias that has gone unexamined.

\begin{figure}[t]
    \centering
    \includegraphics[width=\linewidth]{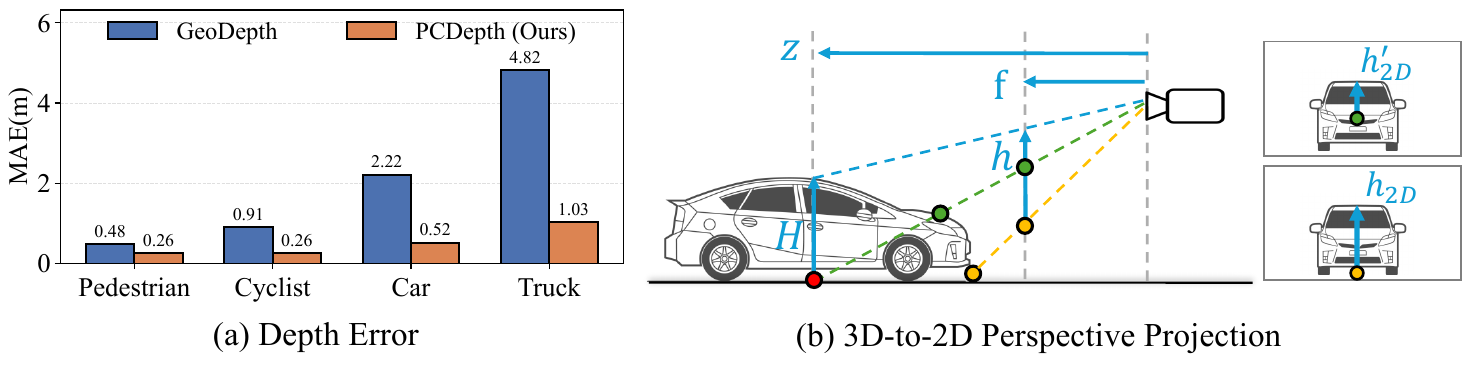}
    \small
    \caption{
        \textbf{(a) Depth error under ideal inputs.}
        We evaluate the GeoDepth formulation~\cite{shi2021geometry} using ground-truth $f$, $H$, and $h_{2D}$.
        Even with these oracle inputs, GeoDepth incurs a non-trivial error that grows with object scale, revealing a structural rather than statistical bias.
        Our \newDepth substantially mitigates this bias across all categories.
        \textbf{(b) Geometric origin of the bias.}
        Under perspective projection, the \threeD bottom center (red) is self-occluded by a nearer surface point (green) that projects higher on the image plane, while the visible bottom in the image corresponds to a still-closer surface point (yellow).
        The commonly adopted formulation $z = f \cdot H / h_{2D}$ therefore recovers the depth of the visible front surface rather than that of the object center, systematically underestimating depth.
    }
    \label{fig:teaser}
    \vspace{-4mm}
\end{figure}

We observe that the formula $z = f \cdot H/h_{2D}$ exhibits a persistent depth error even under \emph{ground-truth} inputs (\cref{fig:teaser}a), and this error grows monotonically with object size, an unmistakable signature of a structural rather than statistical flaw.
Tracing the formula back to its geometric origin reveals the cause.
Under perspective projection, $h_{2D}$ does not encode the depth of the object center but that of the visible front surface.
For any object with non-negligible thickness, the \threeD bottom center is self-occluded by a nearer surface point that projects higher on the image plane (\cref{fig:teaser}b).
The effective projective height tied to the center, $h'_{2D}$, therefore satisfies $h'_{2D} < h_{2D}$, and substituting $h_{2D}$ into the projection relation systematically underestimates center depth.
The bias is small for thin objects and severe for large vehicles, precisely the pattern observed in~\cref{fig:teaser}a.

A naive remedy would be to regress $h'_{2D}$ directly, but $h'_{2D}$ is depth-dependent and not directly observable, making its regression circular and unstable.
We instead retain the measurable $h_{2D}$ and close the gap analytically.
Casting the discrepancy as the object's half-extent along the optical axis, we derive an orientation- and size-aware correction expressible in closed form.
The result is \newDepth, a projection-consistent depth formulation that recovers center depth from $h_{2D}$, the \threeD size, and the yaw angle, eliminating the front-surface bias of geometric estimation (\cref{fig:teaser}a).

A consequence of \newDepth is that depth is no longer an isolated scalar but a function of the predicted \threeD size, orientation, and \twoD height.
This coupling is a double-edged sword: it removes the projection bias but propagates attribute errors into depth, making accurate attribute perception essential.
Lightweight detectors typical of \monoThreeD often underfit the global semantic context required to estimate size and orientation reliably, while simply scaling up the backbone forfeits inference efficiency.
We resolve this tension with \geoDistill, a teacher--student framework in which a high-capacity teacher is built via a Mixture-of-Backbone (\mixtureBackbone) that fuses a convolutional hierarchy with Vision Foundation Models (VFMs), and is then distilled into a standard ResNet student through Hierarchical Feature Alignment (\featureAlignment).
The student inherits the teacher's geometric perception at no additional inference cost, fully exploiting the potential of \newDepth.

In summary, our main contributions are as follows:

$\bullet$ We identify a systematic, projection-induced bias in the conventional geometric depth formulation: substituting the \twoD bounding-box height $h_{2D}$ recovers the depth of the visible front surface rather than the object center, causing size-dependent depth underestimation overlooked by prior work.

$\bullet$ We propose \methodName, a unified framework comprising (1) \newDepth, a projection-consistent depth formulation with orientation- and size-aware corrections that eliminates the projection bias, and (2) \geoDistill, a teacher--student distillation framework that transfers VFM-enriched geometric representations to a lightweight student without inference overhead.

$\bullet$ Extensive experiments demonstrate that \methodName achieves \sota performance on four challenging benchmarks, including a $+1.44\%$ \apThreeD gain on the \kitti leaderboard and $+2.80\%$ on \omnidatasetout, with consistent improvements across object categories and distance ranges.

\section{Related Work}

\noindent\textbf{\monoThreeD.}
Existing \monoThreeD methods fall into two architectural paradigms.
\emph{(1) CNN-based methods}~\cite{ren2016faster} extend \twoD detectors to \threeD using geometric priors and structural constraints~\cite{brazil2019m3d, kumar2021groomed, zhang2021objects, liu2020smoke, ma2021delving, brazil2020kinematic}, with center-based pipelines~\cite{liu2020smoke, ma2021delving, wang2022probabilistic, liu2023monocular} predicting \threeD boxes from image centers and others modeling 2D--3D correlations~\cite{li2022diversity, lu2021geometry, zhang2021objects, wu2024fd3d}.
Complementary lines explore scale-equivariant, camera-aware, and weather-robust designs~\cite{kumar2022deviant, qin2022monoground, zhang2026towards, kumar2025charm3r, jeon2026moca3d, du2026reco, oh2024monowad}, multi-scale FPN structures~\cite{brazil2019m3d, kumar2021groomed, brazil2023omni3d}, BEV segmentation for large objects~\cite{kumar2024seabird}, and auxiliary supervision from CAD models~\cite{chabot2017deep, liu2021autoshape, lee2023baam, krishnan2024omninocs, liu2021voxel}, LiDAR~\cite{ma2019accurate, wang2019pseudo, chen2021monorun, reading2021categorical, peng2024learning, chong2022monodistill, gui2024remote, huang2024training, liu2025monotakd}, synthetic scenes~\cite{lin2025drivegen, lin2024monotta, hu2025adaptive, parihar2025monoplace3d}, and dense depth maps~\cite{ding2020learning, qin2021monogrnet, jiang2024weakly, ma2020rethinking, peng2022did, ranasinghe2024monodiff}, as well as label-efficient training with weak, sparse, or pseudo labels~\cite{zhang2024general, jung2026monosaod, skvrna2025monosowa, lee2026plot, mao2026learnability}; we refer readers to~\cite{ma20233d} for a comprehensive survey.
\emph{(2) Transformer-based methods}~\cite{carion2020end, chen2023group, zhao2024detrs} bring end-to-end detection to \monoThreeD~\cite{huang2022monodtr, wu2023monopgc, zhou2023monoatt, zhang2023monodetr, pu2024monodgp, wang2026span, park2026rare, kim2026probing}, with \monodetr~\cite{zhang2023monodetr} introducing depth-guided queries, \monodgp~\cite{pu2024monodgp} decoupling visual decoding from error-based depth, and \monocop~\cite{zhang2025unleashing} modeling inter-attribute correlations.
\methodName is orthogonal to these architectural advances: rather than redesigning the detector, it targets the depth formulation that all of them share, and can be plugged into both CNN- and transformer-based detectors as a drop-in replacement for the depth head, as we demonstrate in \cref{fig:prodepth-generalization}.

\noIndentHeading{Geometric Depth Estimation in \monoThreeD.}
Geometric projection has long been used to compensate for the lack of metric scale in monocular images, relating observable \twoD cues to \threeD distance through camera intrinsics~\cite{bao2020object, shi2021geometry, barabanau2019monocular, cai2020monocular, lu2021geometry, zhang2023monodetr, pu2024monodgp, yang2025monoclue}. Across these methods, the projection relation $z = fH/h_{2D}$ is adopted as a given, with effort directed at the surrounding pipeline (\eg, keypoint reasoning~\cite{zhang2021objects}, geometric priors~\cite{shi2021geometry}, or uncertainty modeling~\cite{lu2021geometry}) rather than the formula itself. While related lines of work adopt a surface-plus-offset decomposition for regression-based depth~\cite{peng2022did, kim2022boosting} or refine predicted depth post hoc~\cite{zhang2026refineany3d}, the analytical formula and its bias remain unaddressed. We instead identify that substituting the \twoD bounding-box height for the projected center height introduces a systematic projection bias, a structural flaw of the formula that has gone unexamined despite its broad use~\cite{lu2021geometry, zhang2023monodetr, pu2024monodgp, zhang2025unleashing}, and derive an orientation- and size-aware correction directly from perspective geometry.

\noIndentHeading{Knowledge Distillation in \monoThreeD.}
Distillation is a common way to inject depth knowledge into monocular detectors, typically from LiDAR-based teachers whose inputs are unavailable at inference~\cite{chong2022monodistill, liu2025monotakd}. \geoDistill differs in that its teacher consumes the same monocular image as the student, and it transfers VFM-enriched features for attribute perception rather than depth.

\noIndentHeading{Vision Foundation Models.}
Vision Foundation Models (VFMs) learn transferable representations from large-scale vision--language or self-supervised pretraining, with representative families including CLIP~\cite{sun2023eva, zhai2023sigmoid} and DINO~\cite{oquab2023dinov2, simeoni2025dinov3}. While these backbones provide strong semantic priors and have been adapted to \threeD shape understanding~\cite{zhang2024tamm}, their use in geometry-sensitive tasks such as \monoThreeD remains limited~\cite{zhang2026towards, zhang2026refineany3d, yang2025mood}. We leverage VFMs in a teacher--student framework to construct enriched multi-scale representations that support projection-consistent depth modeling under \newDepth.

\section{Method}
\label{sec:method}

\subsection{Problem Setup and Overview}
\label{sec:overview}

\noindent\textbf{Task.}
Given a single RGB image and known camera intrinsics, \monoThreeD aims to detect every object as a \threeD bounding box parameterized by its center $(x, y, z)$, physical size $(W, H, L)$, and yaw angle $\gamma$ in the camera coordinate frame, alongside a class label $C$ and a \twoD bounding box on the image plane.
Among these unknowns, the depth $z$ is the dominant source of error: a small relative error in $z$ produces a metric displacement that scales with depth, making it the central bottleneck of \monoThreeD.

\begin{figure}[t]
    \centering
    \includegraphics[width=0.95\linewidth]{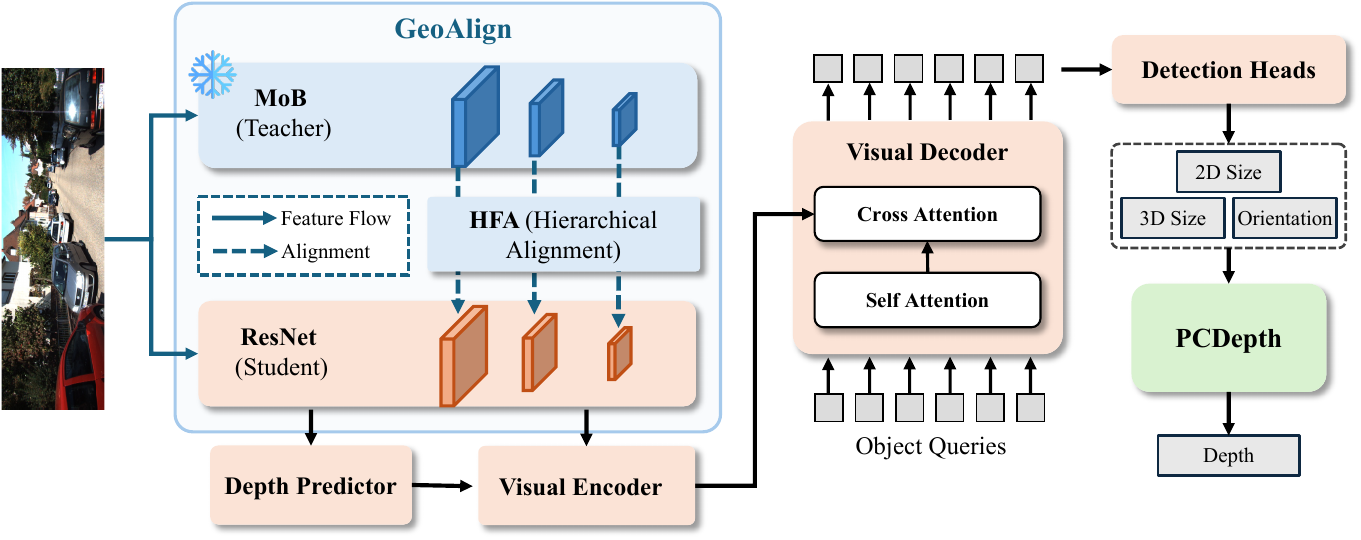}
    \small
    \caption{
        Overview of \methodName.
        The conventional geometric depth $z = f \cdot H / h_{2D}$ recovers the visible front surface rather than the object center, causing a systematic projection bias.
        \newDepth corrects this bias analytically by adding an orientation- and size-aware offset, making depth an explicit function of the predicted 3D size and yaw.
        This shifts the bottleneck onto attribute accuracy, which \geoDistill strengthens by distilling representations from a Mixture-of-Backbone (\mixtureBackbone) teacher into a ResNet student via Hierarchical Feature Alignment (\featureAlignment).
    }
    \label{fig:student-arch}
\end{figure}

\noindent\textbf{Approach at a glance.}
We pursue accurate depth from two angles (see~\cref{fig:student-arch}).
First, we revisit the geometric depth formulation that underpins most modern detectors and identify a structural bias overlooked by prior work (\cref{sec:perspective}).
Second, we correct this bias with \newDepth, a projection-consistent depth formulation that derives center depth from observable \twoD evidence together with predicted \threeD attributes through  geometric correction (\cref{sec:physdepth}).
Because \newDepth couples depth to the predicted size and orientation, depth accuracy now hinges on attribute accuracy.
We close this loop with \geoDistill (\cref{sec:geoDistill}), which distills geometry-aware representations from Vision Foundation Models into a lightweight student, strengthening attribute perception at no inference cost.

\begin{figure}[t]
    \centering
    \includegraphics[width=0.96\linewidth]{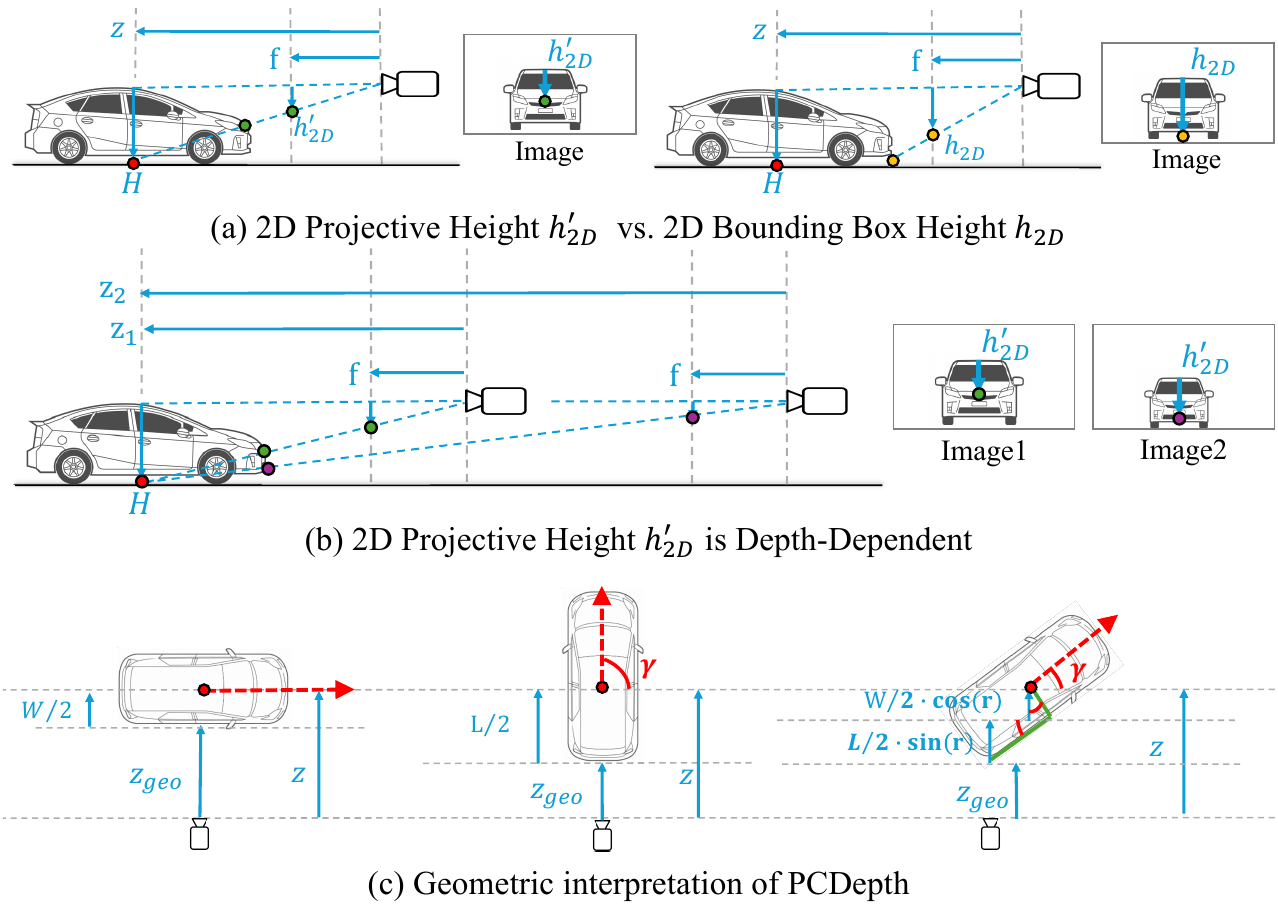}
    \small
    \caption{
        Geometric basis of \newDepth.
        \textbf{(a) Source of the front-surface bias.}
        Under perspective projection, the \threeD bottom center (red) is self-occluded by a nearer surface point (green) that projects higher on the image plane, while the visible bottom of the box corresponds to a still-closer point (yellow), giving $h'_{2D} < h_{2D}$.
        The conventional estimate $z = f \cdot H / h_{2D}$ therefore recovers the front-surface depth rather than the center depth.
        \textbf{(b) $h'_{2D}$ is depth-dependent.}
        As the center depth changes from $z_1$ to $z_2$, the surface point defining $h'_{2D}$ shifts (green to purple), so $h'_{2D}$ is not directly observable and its regression is ill-posed.
        \textbf{(c) Recovering center depth via an optical-axis correction.}
        $z_{\text{geo}}$ recovers the front-surface depth, displaced from the object center by an offset $\Delta$ along the optical axis: $\Delta \approx W/2$ at $\gamma = 0^{\circ}$ (\textbf{Left}), $\Delta \approx L/2$ at $\gamma = 90^{\circ}$ (\textbf{Middle}), and $\Delta \approx \tfrac{L}{2}|\sin\gamma| + \tfrac{W}{2}|\cos\gamma|$ for arbitrary yaw (\textbf{Right}), giving $z = z_{\text{geo}} + \Delta$.
    }
    \label{fig:geo}
\end{figure}

\subsection{The Front-Surface Bias of Geometric Depth}
\label{sec:perspective}

The geometric depth formulation traces back to the pinhole camera model, under which the depth of a point at \threeD height $H$ projecting to image height $h$ is
\begin{equation}
z = f \cdot \frac{H}{h},
\label{eq:geo}
\end{equation}
with $f$ the focal length.
In the \monoThreeD literature~\cite{zhang2023monodetr,lu2021geometry,pu2024monodgp}, $h$ is instantiated by the predicted \twoD bounding-box height $h_{2D}$, yielding $z_{\text{geo}} = f \cdot H / h_{2D}$.
This substitution is intuitive but, as we now show (\cref{fig:geo}a), geometrically inconsistent.

\noindent\textbf{Diagnosis.}
\cref{eq:geo} is exact only when $h$ is the projection of the line segment passing through the object center perpendicular to the optical axis.
For an object with non-negligible thickness, this center-aligned segment is \emph{not} what the \twoD bounding box captures.
As shown in~\cref{fig:geo}a, the \threeD bottom center (red) is self-occluded by a nearer surface point (green) that projects higher on the image plane; the visible bottom of the box corresponds to a still-closer surface point (yellow).
The effective center-aligned projective height $h'_{2D}$ therefore satisfies $h'_{2D} < h_{2D}$, and $z_{\text{geo}}$ recovers the depth of the visible front surface rather than that of the object center.
We refer to the resulting offset as the \emph{front-surface bias}; \cref{fig:teaser}a shows that it grows monotonically with object size and persists even under ground-truth $f$, $H$, and $h_{2D}$, confirming a structural rather than statistical origin.

\subsection{\newDepth: A Projection-Consistent Reformulation}
\label{sec:physdepth}

\noindent\textbf{Why not regress }$\mathbf{h'_{2D}}$\textbf{?}
The most direct fix is to predict $h'_{2D}$ and substitute it into \cref{eq:geo}.
This is, however, ill-posed: $h'_{2D}$ is not a property that can be read off the image but is itself a function of the depth we seek,
\begin{equation}
h'_{2D} = g(z;\, \gamma, \mathbf{s}),
\label{eq:hprime}
\end{equation}
where $\mathbf{s}=(W, H, L)$ is the \threeD size.
\cref{fig:geo}b illustrates this dependency: changing $z$ shifts the visible surface point that defines $h'_{2D}$, so any regressor for $h'_{2D}$ is implicitly conditioned on the very depth it is meant to estimate, creating a circular dependency that destabilizes optimization.

\noindent\textbf{What does} $\mathbf{h_{2D}}$ \textbf{actually encode?}
We instead retain the directly measurable $h_{2D}$ and derive its geometric meaning from first principles.
Under the pinhole model, the eight corners of the \threeD bounding box project to image points $\bigl(fX_i/Z_i,\, fY_i/Z_i\bigr)$, and $h_{2D}$ is the spread of their vertical projections.
For an upright box with yaw-only orientation, as in standard driving benchmarks, all four top corners share $Y = -H/2$ and all four bottom corners share $Y = +H/2$, so the extremal image-plane $y$-values are both attained at the corner with the \emph{minimum} depth $Z_{\min}$. This gives
\begin{equation}
h_{2D} \;=\; \frac{f H}{Z_{\min}},
\qquad
Z_{\min} \;=\; z - \Delta,
\label{eq:hbox}
\end{equation}
where $z$ is the center depth and $\Delta$ is the optical-axis half-extent of the box footprint.
Substituting into the conventional formula makes the front-surface bias explicit:
\begin{equation}
z_{\text{geo}} \;=\; f \cdot \frac{H}{h_{2D}} \;=\; Z_{\min} \;=\; z - \Delta,
\qquad\text{i.e.,}\qquad
z \;=\; z_{\text{geo}} + \Delta.
\label{eq:correction}
\end{equation}
$z_{\text{geo}}$ recovers the depth of the box's nearest corner along the optical axis rather than that of the center, and the gap is exactly $\Delta$.

\noindent\textbf{Closed-form correction.}
For a box yawed by $\gamma$, the four bottom-footprint corners have depths $z \pm (L/2)\sin\gamma \pm (W/2)\cos\gamma$, so the minimum is attained at the corner closest to the camera:
\begin{equation}
\Delta \;=\; \frac{L}{2}|\sin\gamma| + \frac{W}{2}|\cos\gamma|,
\label{eq:delta}
\end{equation}
which reduces to $\Delta = W/2$ at $\gamma = 0^{\circ}$ (object facing the camera) and $\Delta = L/2$ at $\gamma = 90^{\circ}$ (object sideways), matching the limit cases in \cref{fig:geo}c.
Combining \cref{eq:correction} and \cref{eq:delta} yields \newDepth:
\begin{equation}
z \;=\; f \cdot \frac{H}{h_{2D}}
\;+\; \frac{L}{2}|\sin\gamma|
\;+\; \frac{W}{2}|\cos\gamma|.
\label{eq:prodepth}
\end{equation}
Under pinhole projection and an upright-box orientation, this identity holds analytically when $h_{2D}$ is the tight projection of the eight 3D corners. In real datasets, $h_{2D}$ may deviate from this tight projection due to truncation, occlusion, or annotation noise, but PCDepth still provides a stronger geometric guarantee than regression-based formulations, as confirmed in \cref{fig:teaser}a.

\begin{figure}[t]
    \centering
    \includegraphics[width=\linewidth]{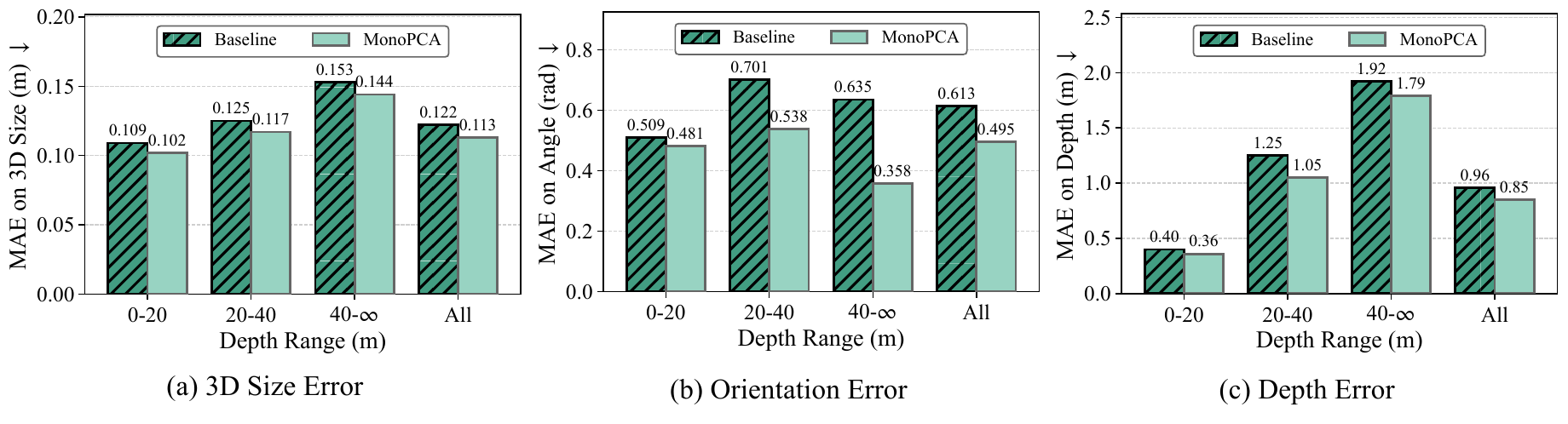}
    \small
    \caption{
        \geoDistill reduces depth error through the attributes \newDepth depends on.
        Per-distance errors on \kitti \val for \threeD size \textbf{(Left)}, orientation \textbf{(Middle)}, and depth \textbf{(Right)}. \geoDistill consistently lowers size and orientation errors across all ranges, and since \newDepth computes depth analytically from these two attributes, the improvements propagate directly to depth.
    }
    \label{fig:student-attribute}
    \vspace{-2mm}
\end{figure}

\subsection{\geoDistill: Strengthening Attribute Perception}
\label{sec:geoDistill}
\newDepth ties depth accuracy to the predicted size and yaw, shifting the bottleneck to attribute perception.
Both attributes are difficult to recover from a single image: size requires category-level priors inferred without metric scale, and yaw relies on subtle appearance cues such as part configuration and self-shading.
Such cues call for representations that combine fine local geometry with long-range semantic context, a regime where Vision Foundation Models (VFMs) excel, yet naive backbone replacement would multiply inference cost.
\geoDistill resolves this tension by distilling a high-capacity teacher, formed by fusing a VFM with the convolutional backbone, into a lightweight student, transferring attribute perception (\cref{fig:student-attribute}) without inference overhead.

\begin{wrapfigure}{r}{0.55\linewidth}
    \centering
    \vspace{-1mm}
    \includegraphics[width=\linewidth]{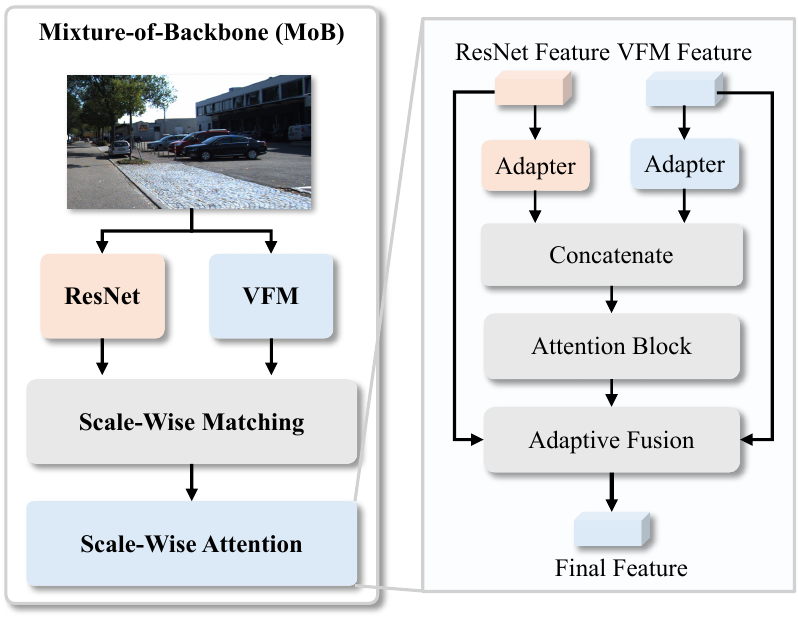}
    \vspace{-4mm}
    \small
    \caption{
        Details of Mixture-of-Backbone.
        \mixtureBackbone adaptively fuses a convolutional pyramid with a single-scale VFM branch at every scale.
    }
    \label{fig:mob}
    \vspace{-6mm}
\end{wrapfigure}
\noindent\textbf{Mixture-of-Backbone.}
The two branches are structurally incompatible out of the box: the convolutional backbone produces a four-level pyramid at strides $\{1/8, 1/16, 1/32, 1/64\}$, with the last level obtained by a stride-2 convolution on the $1/32$ map, while the VFM emits a single-scale representation.
\mixtureBackbone reconciles them in two steps (\cref{fig:mob}).

\textit{Scale-Wise Matching} expands the VFM output into a four-level pyramid through an FPN~\cite{li2022exploring}, producing $F^{V}_{s}$ at each scale $s$ spatially aligned with the convolutional features $F^{R}_{s}$.

\textit{Scale-Wise Attention} fuses the aligned features adaptively.
Lightweight adapters $\phi_R, \phi_V$ map both streams into a shared embedding space, $\tilde{F}^{R}_{s} = \phi_R(F^{R}_{s})$ and $\tilde{F}^{V}_{s} = \phi_V(F^{V}_{s})$.
A small attention module then produces a per-location weight map and the fused teacher feature:
\begin{equation}
A_s = \sigma\bigl(\mathcal{A}([\tilde{F}^{R}_{s}, \tilde{F}^{V}_{s}])\bigr),
\qquad
F^{T}_{s} = A_s \odot \tilde{F}^{R}_{s} + (1 - A_s) \odot \tilde{F}^{V}_{s},
\end{equation}
with $\sigma$ the element-wise sigmoid.
By learning $A_s$ per location, \mixtureBackbone defers to convolutional features where local geometric detail dominates and to the VFM where semantic context dominates.

\noindent\textbf{Hierarchical Feature Alignment.}
Given the teacher features $\{F^{T}_{s}\}_{s=1}^{4}$ and student features $\{F^{S}_{s}\}_{s=1}^{4}$ from a standard ResNet, \featureAlignment aligns the two patch-wise across all scales.
We project the student into the teacher's embedding space via $\hat{F}^{S}_{s} = \text{Proj}(F^{S}_{s})$ and minimize the patch-wise cosine distance:
\begin{equation}
\mathcal{L}_{\text{align}}
= \sum_{s=1}^{4}
\frac{1}{N_s}
\sum_{i=1}^{N_s}
\left(
1 -
\frac{
\langle \hat{F}^{S}_{s,i}, F^{T}_{s,i} \rangle
}{
\|\hat{F}^{S}_{s,i}\|_2 \, \|F^{T}_{s,i}\|_2
}
\right),
\end{equation}
where $i$ indexes spatial locations and $N_s$ the number of patches at scale $s$.
Cosine is invariant to the VFM-induced magnitude gap between teacher and student, and we span all four scales to capture attribute cues at multiple resolutions.

\noindent\textbf{Overall Loss.}
The training objective combines the standard DETR-style detection losses with the alignment term, $\mathcal{L} = \mathcal{L}_{\text{det}} + \lambda_{\text{align}}\,\mathcal{L}_{\text{align}}$, where $\mathcal{L}_{\text{det}}$ supervises Hungarian-matched predictions with classification, \twoD box, projected-center, depth, size, orientation, and depth-map losses, and $\lambda_{\text{align}}$ balances the two terms.
The individual losses and their weights are given in Appendix~\ref{appendix-imple}.

\section{Experiments}
\label{sec:exp}

\noindent\textbf{Datasets.} We evaluate \methodName on four representative \monoThreeD benchmarks.
\textbf{\kitti}~\cite{geiger2012we} contains $7{,}481$ training and $7{,}518$ testing images; following~\cite{chen2016monocular}, the training split is further divided into $3{,}712$ training and $3{,}769$ validation images.
\textbf{\nuscenes}~\cite{caesar2020nuscenes} provides $28{,}130$ training and $6{,}019$ validation front-camera images, with annotations converted to \kitti format following~\cite{kumar2022deviant}.
\textbf{\waymo}~\cite{sun2020scalability} provides $52{,}386$ training and $39{,}848$ validation front-view samples~\cite{kumar2022deviant}.
\textbf{\omnidataset}~\cite{brazil2023omni3d} aggregates multiple \threeD detection datasets into a unified benchmark; we evaluate on \omnidatasetout, \omnidatasetkitti, and \omnidatasetnu, and additionally report \omnidatasetoutsmall, \omnidatasetoutlarge, and \omnidatasetoutcar splits to disentangle performance across object scales.

\noindent\textbf{Evaluation Metrics.}
For \kitti and \nuscenes, we follow~\cite{simonelli2019disentangling} and report Car \apThreeD and \apBev at \iouThreeD $\ge 0.7$, with Moderate as the primary setting, and additionally report \iouThreeD $\ge 0.5$ on \nuscenes. For \waymo, we report \apThreeD and the heading-aware \aphThreeD~\cite{reading2021categorical, kumar2022deviant} at \iouThreeD $\ge 0.5$ and $0.7$ over distance ranges $[0,30)$, $[30,50)$, and $[50,\infty)$ meters. For \omnidataset, we report AP averaged over \iouThreeD thresholds $\tau \in [0.05, 0.10, \dots, 0.50]$.

\noindent\textbf{Implementation Details.}
We build \methodName on \monocop~\cite{zhang2025unleashing}. The student uses a ResNet-50 backbone, while the teacher fuses ResNet-50 with EVA-CLIP~\cite{sun2023eva} through \mixtureBackbone and is discarded after training. We use AdamW with an initial learning rate of $2\times10^{-4}$ ($2\times10^{-5}$ for the teacher backbone) and a weight decay of $10^{-4}$, and halve the learning rate at epochs 85, 125, 165, and 205. The alignment loss weight is $\lambda_{align}=5$ (Appendix~\ref{appendix-ablation-alignment-loss}). All models are trained with a batch size of 16 on eight NVIDIA A100 GPUs; a full \kitti run takes about 12 hours. The full architecture, training losses, and loss weights are given in Appendix~\ref{appendix-imple}.

\begin{table*}[t]
    \centering
    \small
    \caption{
        \kitti Leaderboard (\test) and \val results at \iouThreeD $\ge 0.7$.  Without extra data, \methodName achieves \sota performance across all metrics on both the \kitti leaderboard (Test) and \kitti val.  [Key: \textbf{Best}, \underline{Second Best}, \textcolor{myblue}{\textbf{Gain}}]
    }
    \tabcolsep=0.1cm
    \resizebox{1\textwidth}{!}{
        \begin{tabular}{l|c|ccc|ccc|ccc|ccc}
            \midrule
            \multirow{2}{*}{Method} & \multirow{2}{*} {\makecell{Extra \\ Data}} & \multicolumn{3}{c|}{ \textbf{\test}, \apThreeD $\uparrow$ } & \multicolumn{3}{c|}{ \textbf{\test}, \apBev $\uparrow$ } & \multicolumn{3}{c|}{ \textbf{\val}, \apThreeD $\uparrow$ } & \multicolumn{3}{c}{ \textbf{\val}, \apBev $\uparrow$ } \\
                                                      &       & Easy  & Mod.  & Hard  & Easy      & Mod.      & Hard      & Easy  & Mod.  & Hard  & Easy      & Mod.      & Hard \\
            \midrule
            \occupancymThreeD~\cite{peng2024learning} & LiDAR & 25.55 & 17.02 & 14.79 & 35.38     & 24.18     & 21.37     & 26.87 & 19.96 & 17.15 & 35.72     & 26.60     & 23.68 \\
            \opaThreeD~\cite{su2023opa}               & Depth & 24.68 & 17.17 & 14.14 & 32.50     & 23.14     & 20.30     & 24.97 & 19.40 & 16.59 & 33.80     & 25.51     & 22.13 \\
            \monotakd~\cite{liu2025monotakd}          & LiDAR & 27.91 & 19.43 & 16.51 & 38.75     & 27.76     & 24.14     & 34.36 & 22.61 & 19.88 & 42.86     & 29.41     & 26.47 \\
            \midrule
            \monocon{}~\cite{liu2022monocon}          & \multirow{9}{*}{None} & 22.50 & 16.46 & 13.95 & 31.12     & 22.10     & 19.00     & 26.33 & 19.01 & 15.98 & \mathDash & \mathDash & \mathDash \\
            \monodetr~\cite{zhang2023monodetr}        &       & 25.00 & 16.47 & 13.58 & 33.60     & 22.11     & 18.60     & 28.84 & 20.61 & 16.38 & 37.86     & 26.95     & 22.80 \\
            \monouni~\cite{jia2023monouni}            &       & 24.75 & 16.73 & 13.49 & \mathDash & \mathDash & \mathDash & 24.51 & 17.18 & 14.01 & \mathDash & \mathDash & \mathDash \\
            \fdThreeD~\cite{wu2024fd3d}               &       & 25.38 & 17.12 & 14.50 & 34.20     & 23.72     & 20.76     & 28.22 & 20.23 & 17.04 & 36.98     & 26.77     & 23.16 \\
            \monomae~\cite{jiang2024monomae}          &       & 25.60 & 18.84 & \underline{16.78} & 34.15     & 24.93     & 21.76     & 30.29 & 20.90 & 17.61 & 40.26     & 27.08     & 23.14 \\
            \monocd~\cite{yan2024monocd}              &       & 25.53 & 16.59 & 14.53 & 33.41     & 22.81     & 19.57     & 26.45 & 19.37 & 16.38 & 34.60     & 24.96     & 21.51 \\
            \monodgp~\cite{pu2024monodgp}             &       & 26.35 & 18.72 & 15.97 & 35.24     & 25.23     & 22.02     & 30.76 & 22.34 & 19.02 & 39.40     & 28.20     & 24.42 \\
            \monocop~\cite{zhang2025unleashing}       &       & \underline{27.54} & \underline{19.11} & 16.33 & \underline{36.77} & \underline{25.57} & \underline{22.62} & \underline{32.06} & \underline{23.98} & \underline{20.64} & \underline{41.92} & \underline{30.75} & \underline{27.06} \\
            \textbf{\methodName}                      &       & \textbf{30.34} & \textbf{20.55} & \textbf{17.84} & \textbf{38.65} & \textbf{26.63} & \textbf{24.43} & \textbf{35.86} & \textbf{26.22} & \textbf{23.58} & \textbf{46.29} & \textbf{33.89} & \textbf{30.05} \\
            \rowcolor{hdColor}
            \textcolor{myblue}{\textbf{$\Delta$}}     &       & \textcolor{myblue}{\textbf{+2.80}} & \textcolor{myblue}{\textbf{+1.44}} & \textcolor{myblue}{\textbf{+1.06}} & \textcolor{myblue}{\textbf{+1.88}} & \textcolor{myblue}{\textbf{+1.06}} & \textcolor{myblue}{\textbf{+1.81}} & \textcolor{myblue}{\textbf{+3.80}} & \textcolor{myblue}{\textbf{+2.24}} & \textcolor{myblue}{\textbf{+2.94}} & \textcolor{myblue}{\textbf{+4.37}} & \textcolor{myblue}{\textbf{+3.14}} & \textcolor{myblue}{\textbf{+2.99}} \\
            \midrule
        \end{tabular}}
    \label{exp:kitti-test-val}
    \vspace{-2mm}
\end{table*}

\begin{table}[t]
    \centering
    \caption{
        \nuscenes \val results. \methodName achieves \sota performance on both \threeD and \bev detection under IoU thresholds 0.7 and 0.5.
        [Key: \textbf{Best}, \underline{Second Best}, \textcolor{myblue}{\textbf{Gain}}]
    }
    \tabcolsep=0.25cm
    \resizebox{0.8\textwidth}{!}{
        \begin{tabular}{l|cc|cc|cc|cc}
            \midrule
            \multirow{3}{*}{Method} & \multicolumn{4}{c|}{\textbf{\iouThreeD $\ge$ 0.7}} & \multicolumn{4}{c}{\textbf{\iouThreeD $\ge$ 0.5}} \\
            \cmidrule(lr){2-5} \cmidrule(lr){6-9}
            & \multicolumn{2}{c|}{\apThreeD $\uparrow$} & \multicolumn{2}{c|}{\apBev $\uparrow$} & \multicolumn{2}{c|}{\apThreeD $\uparrow$} & \multicolumn{2}{c}{\apBev $\uparrow$} \\
            \cmidrule(lr){2-3} \cmidrule(lr){4-5} \cmidrule(lr){6-7} \cmidrule(lr){8-9}
                                                      & Easy  & Mod. & Easy  & Mod.  & Easy  & Mod.  & Easy  & Mod. \\
            \midrule
            \gupNet~\cite{lu2021geometry}             & 8.50  & 7.40 & 14.21 & 12.81 & 29.03 & 26.16 & 33.42 & 30.23 \\
            \deviant~\cite{kumar2022deviant}          & 9.69  & 8.33 & 16.28 & 14.36 & 31.47 & 28.22 & 35.61 & 31.93 \\
            \monodetr~\cite{zhang2023monodetr}        & 9.53  & 8.19 & 16.39 & 14.41 & 31.81 & 28.35 & 35.70 & 31.96 \\
            \monodgp~\cite{pu2024monodgp}             & 10.04 & 8.78 & 16.55 & 14.53 & 29.56 & 26.17 & 32.67 & 29.44 \\
            \monocop~\cite{zhang2025unleashing}       & \underline{10.85} & \underline{9.71} & \underline{17.83} & \underline{15.86} & \underline{33.70} & \underline{29.91} & \underline{37.44} & \underline{34.01} \\
            \textbf{\methodName}                      & \textbf{13.75} & \textbf{12.03} & \textbf{20.97} & \textbf{18.49} & \textbf{38.50} & \textbf{34.11} & \textbf{42.03} & \textbf{38.49} \\
            \rowcolor{hdColor}
            \textcolor{myblue}{\textbf{$\Delta$}}     & \textcolor{myblue}{\textbf{+2.90}} & \textcolor{myblue}{\textbf{+2.32}} & \textcolor{myblue}{\textbf{+3.14}} & \textcolor{myblue}{\textbf{+2.63}} & \textcolor{myblue}{\textbf{+4.80}} & \textcolor{myblue}{\textbf{+4.20}} & \textcolor{myblue}{\textbf{+4.59}} & \textcolor{myblue}{\textbf{+4.48}} \\
            \midrule
        \end{tabular}
    }

    \label{exp:nuscenes-val-main}
    \vspace{-2mm}
\end{table}

\subsection{Standard \threeD Benchmarks}
We evaluate \methodName on three standard \monoThreeD benchmarks (\kitti~\cite{geiger2012we}, \nuscenes~\cite{caesar2020nuscenes}, and \waymo~\cite{sun2020scalability}), where each model is trained and tested on a single dataset.

\noindent\textbf{\kitti Leaderboard (\test) Results.}
The \kitti leaderboard withholds ground-truth annotations and evaluates submissions on the official server, providing a fully blind evaluation protocol. As shown in \cref{exp:kitti-test-val}, \methodName achieves new \sota performance across all \threeD and \bev metrics without any extra supervision. Under the primary \apThreeD Mod. metric, \methodName reaches $20.55\%$, outperforming \monocop by $+1.44\%$. The improvement holds across all difficulty levels and extends to \bev evaluation, indicating more accurate \threeD localization. \methodName also surpasses methods that rely on additional supervision such as LiDAR or depth maps (\eg, MonoTAKD and OPA-3D) on all \apThreeD metrics. More results are provided in Appendix~\ref{appendix-kitti}.

\noindent\textbf{\kitti \val Results.}
We observe similar trends on the \val split. \methodName achieves $26.22\%$ under \apThreeD Mod., exceeding \monocop by $+2.24\%$. The gains hold across all difficulty levels, including the Hard setting dominated by occluded and distant objects, suggesting that they stem from better geometric depth rather than from a subset of easy cases.

\noindent\textbf{\nuscenes Val Results.}
As shown in \cref{exp:nuscenes-val-main}, the gains carry over to \nuscenes: \methodName leads on every \threeD and \bev metric and improves \apThreeD Mod. by $+2.32\%$ over \monocop at IoU $\ge 0.7$. The margin widens at IoU $\ge 0.5$ ($+4.20\%$), suggesting that \newDepth removes a systematic depth offset that pushes predictions out of the matching range, rather than only tightening well-localized boxes.

\noindent\textbf{\waymo Val Results.}
On \waymo (Appendix~\ref{appendix-waymo}), \methodName achieves the best overall \apThreeD and \aphThreeD at IoU $\ge 0.5$, with gains concentrated within $50$\,m ($+4.77\%$ and $+1.61\%$ \apThreeD over \monocop at $0$--$30$\,m and $30$--$50$\,m). At these distances depth noise is small and the front-surface offset is a dominant error source, so correcting it pays off most; farther away, noise that grows with depth limits the benefit of any correction.

\begin{wraptable}{r}{0.5\linewidth}
    \centering
    \vspace{-2mm}
    \small
    \caption{Efficiency comparison.}
    \vspace{-2mm}
    \tabcolsep=0.12cm
    \resizebox{1\linewidth}{!}{
        \begin{tabular}{lccc}
            \midrule
            Method                              & AP$_{3D} \uparrow$ & Param(M) $\downarrow$ & GFLOPs(G) $\downarrow$ \\
            \midrule
            \monodgp~\cite{pu2024monodgp}       & 18.72              & 38.90                 & 69.0 \\
            \monocop~\cite{zhang2025unleashing} & 19.11              & 42.50                 & 71.8 \\
            \textbf{\methodName}                & \textbf{20.55}     & 42.00                 & 62.5 \\
            \midrule
        \end{tabular}}
    \vspace{-6mm}
    \label{tab:eff}
\end{wraptable}
\noindent\textbf{Efficiency Analysis.}
Beyond accuracy, \cref{tab:eff} shows that \methodName is also more efficient on the \kitti leaderboard: AP$_{3D}$ improves from $19.11\%$ (\monocop) to $20.55\%$, while GFLOPs drop from $71.8$ to $62.5$ at comparable parameters. The GFLOPs reduction reflects an architectural simplification enabled by PCDepth: computing depth in closed form from already-predicted attributes removes the need for a dedicated depth encoder and depth-guided decoder. PCDepth is parameter-free and the GeoAlign teacher is discarded after training, so neither adds inference cost.

\subsection{\omnidataset Benchmark}
Beyond single-dataset evaluations, we assess \methodName on the large-scale \omnidataset benchmark, which jointly trains and evaluates on multiple datasets. This setting requires consistent performance across domains rather than overfitting to a single dataset.
As shown in \cref{exp:omni3d}, \methodName achieves new \sota performance on the unified \omnidatasetout split, reaching $35.0\%$ AP and outperforming the recent foundation-based DetAny3D by $+2.8\%$. It also surpasses the best prior method on both the \omnidatasetkitti and \omnidatasetnu subsets ($+2.1\%$ and $+3.1\%$), indicating that the improvements do not stem from dataset-specific tuning. At the category level, \methodName obtains $+4.1\%$ on \textit{Car}, the dominant class in outdoor scenes, and $+3.6\%$ on small objects. Per-category results are provided in Appendix~\ref{appendix-omni3d}.
Together with the \kitti leaderboard results above, \methodName is, to the best of our knowledge, the first monocular framework to reach \sota on both benchmarks.

\begin{table}[t]
    \centering
    \caption{
        \omnidataset results.  To the best of our knowledge, \methodName is \textbf{the first model} that achieves both \sota performance under \omnidatasetout benchmark and \kitti leaderboard, showcasing its superior \threeD detection performance. [Key: \textbf{Best}, \underline{Second Best}, \textcolor{myblue}{\textbf{Gain}}]}
    \tabcolsep=0.15cm
    \resizebox{1\linewidth}{!}{
        \begin{tabular}{l|cccccc|ccc}
            \midrule
            \multirow{2}{*}{Method} & \multicolumn{6}{c|}{\omnidatasetout} & \multicolumn{3}{c}{\rm \kitti Leaderboard} \\
                                                      & ${\rm AP^{kit}_{3D}} \uparrow$ & ${\rm AP^{nus}_{3D}} \uparrow$ & ${\rm AP^{out}_{3D}} \uparrow$ & ${\rm AP^{large}_{3D}} \uparrow$ & ${\rm AP^{small}_{3D}} \uparrow$ & ${\rm AP^{car}_{3D}} \uparrow$ & Easy$\uparrow$ & Mod.$\uparrow$ & Hard$\uparrow$ \\
            \midrule
            \monodgp~\cite{pu2024monodgp}             & \mathDash                      & \mathDash                      & \mathDash                      & \mathDash                        & \mathDash                        & \mathDash                      & 26.35          & 18.72          & 15.97 \\
            \monocop~\cite{zhang2025unleashing}       & \mathDash                      & \mathDash                      & \mathDash                      & \mathDash                        & \mathDash                        & \mathDash                      & \underline{27.54} & \underline{19.11} & \underline{16.33} \\
            ImVoxelNet~\cite{rukhovich2022imvoxelnet} & 23.5                           & 23.4                           & 21.5                           & \mathDash                        & \mathDash                        & \mathDash                      & 17.15          & 10.97          & 9.15 \\
            SMOKE~\cite{liu2020smoke}                 & 25.9                           & 20.4                           & 20.0                           & \mathDash                        & \mathDash                        & \mathDash                      & 14.03          & 9.76           & 7.84 \\
            Cube R-CNN~\cite{brazil2023omni3d}        & \underline{36.0}               & 32.7                           & 31.9                           & \underline{37.6}                 & \underline{21.8}                 & \underline{72.9}               & 23.59          & 15.01          & 12.56 \\
            DetAny3D~\cite{zhang2025detect}           & 35.8                           & \underline{33.9}               & \underline{32.2}               & \mathDash                        & \mathDash                        & \mathDash                      & 26.89          & 18.67          & 15.48 \\
            \textbf{\methodName}                      & \textbf{38.1}                  & \textbf{37.0}                  & \textbf{35.0}                  & \textbf{38.8}                    & \textbf{25.4}                    & \textbf{77.0}                  & \textbf{30.34} & \textbf{20.55} & \textbf{17.84} \\
            \rowcolor{hdColor}
            \textcolor{myblue}{\textbf{$\Delta$}}     & \textcolor{myblue}{\textbf{+2.1}} & \textcolor{myblue}{\textbf{+3.1}} & \textcolor{myblue}{\textbf{+2.8}} & \textcolor{myblue}{\textbf{+1.2}} & \textcolor{myblue}{\textbf{+3.6}} & \textcolor{myblue}{\textbf{+4.1}} & \textcolor{myblue}{\textbf{+2.80}} & \textcolor{myblue}{\textbf{+1.44}} & \textcolor{myblue}{\textbf{+1.51}} \\
            \midrule
        \end{tabular}
    }

    \label{exp:omni3d}
    \vspace{-2mm}
\end{table}

\begin{figure}[t]
    \centering
    \includegraphics[width=\linewidth]{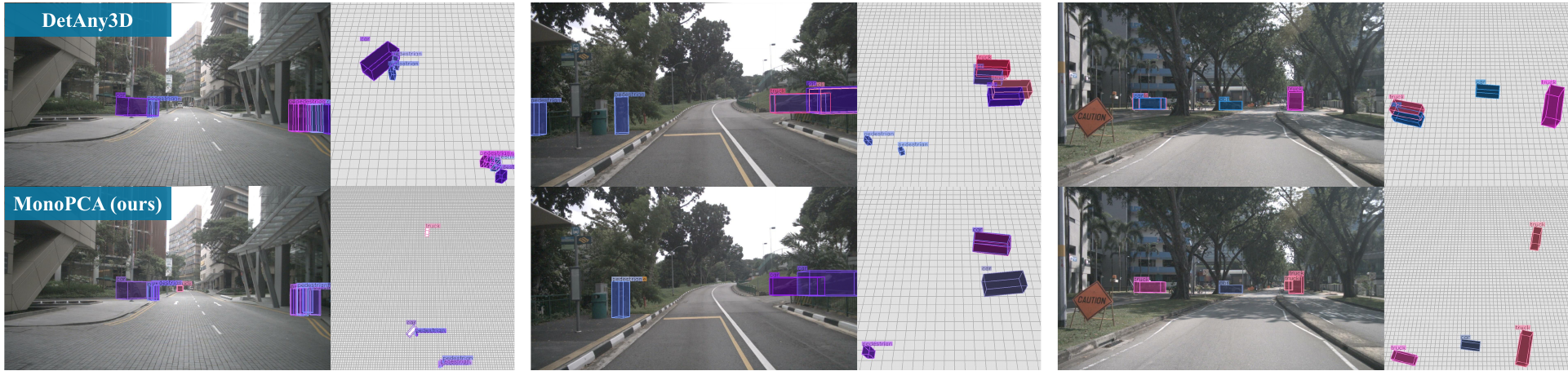}
    \caption{
        Qualitative comparison on \omnidatasetout. \threeD predictions overlaid on the image (Left) and the corresponding bird's-eye view with a $1$m$\times$$1$m grid (Right). }
    \label{fig:qualitative}
    \vspace{-2mm}
\end{figure}

\noindent\textbf{Qualitative Analysis.}
We further provide qualitative detection results in~\cref{fig:qualitative}.
Compared with DetAny3D~\cite{zhang2025detect}, \methodName produces more accurate detections for distant objects as well as small objects such as pedestrians.
Additional visualization results are in~\cref{appendix-vis}.

\subsection{Ablation Study}

We conduct ablation studies on the \kitti \val split, reporting \apThreeD under \iouThreeD\,$\geq$\,$0.7$ with the Mod. setting as the primary metric. Additional results are provided in Appendix~\ref{appendix-ablation}.

\noindent\textbf{Component Analysis.}
\cref{tab:ablation-all}a reports the contribution of each proposed component. \newDepth and \geoDistill alone improve the baseline by $+1.09\%$ and $+0.75\%$, respectively. Combined, they yield a $+2.24\%$ gain, exceeding the sum of the individual gains ($+1.84\%$). The two components are therefore complementary by design: \newDepth couples depth to the predicted size and orientation, making attribute accuracy the new bottleneck, and \geoDistill addresses this bottleneck by distilling VFM-enriched representations into the student.

\noindent\textbf{\newDepth Design.}
\cref{tab:ablation-all}b compares different depth formulations. Direct regression achieves $23.82\%$, indicating that depth prediction without geometric grounding is suboptimal. \geoDepth with the measurable $h_{2D}$ improves the result to $24.73\%$, validating projection-based modeling, but inherits the front-surface bias since $h_{2D}$ encodes the projected front face rather than the object center. Replacing $h_{2D}$ with the bias-free $h'_{2D}$ drops performance to $24.18\%$, because $h'_{2D}$ is depth-dependent and not directly observable, making its regression unstable. \newDepth retains the measurable $h_{2D}$ and applies an orientation- and size-aware correction, preserving observability while restoring geometric consistency, and achieves the best result of $26.22\%$ ($+1.49\%$ over \geoDepth).

\noindent\textbf{\mixtureBackbone Design.}
\cref{tab:ablation-all}c evaluates the design choice of \mixtureBackbone. Removing \scaleMatch drops performance to $24.97\%$ ($-1.25\%$), while removing \scaleAttention drops it to $25.44\%$ ($-0.78\%$). \scaleMatch has a larger impact because the VFM produces only a single-scale feature; without an FPN-style expansion, it cannot be aligned with the convolutional pyramid at strides $\{1/8, 1/16, 1/32, 1/64\}$ for scale-wise fusion. \mixtureBackbone (full) at $26.22\%$ also outperforms a VFM-only encoder ($25.78\%$) by $+0.44\%$ and the ResNet-50 backbone ($25.07\%$) by $+1.15\%$, indicating that convolutional features and VFM features carry complementary information rather than redundant capacity.

\noindent\textbf{Alignment Loss Design.}
\cref{tab:ablation-all}d compares alignment objectives. Both MAE ($25.53\%$) and MSE ($25.69\%$) improve over the no-distillation baseline ($25.07\%$), but \featureAlignment achieves the best result of $26.22\%$. We attribute the gap to feature-magnitude mismatch: the teacher includes VFM-derived components whose magnitudes a ResNet student cannot reproduce, so element-wise regression losses pull the student toward an unattainable scale. \featureAlignment uses cosine similarity, which is invariant to magnitude and aligns feature directions patch-wise across all four scales, making it suitable for distillation between heterogeneous backbones.

\begin{wrapfigure}{r}{0.4\linewidth}
    \centering
    \vspace{-2mm}
    \includegraphics[width=\linewidth]{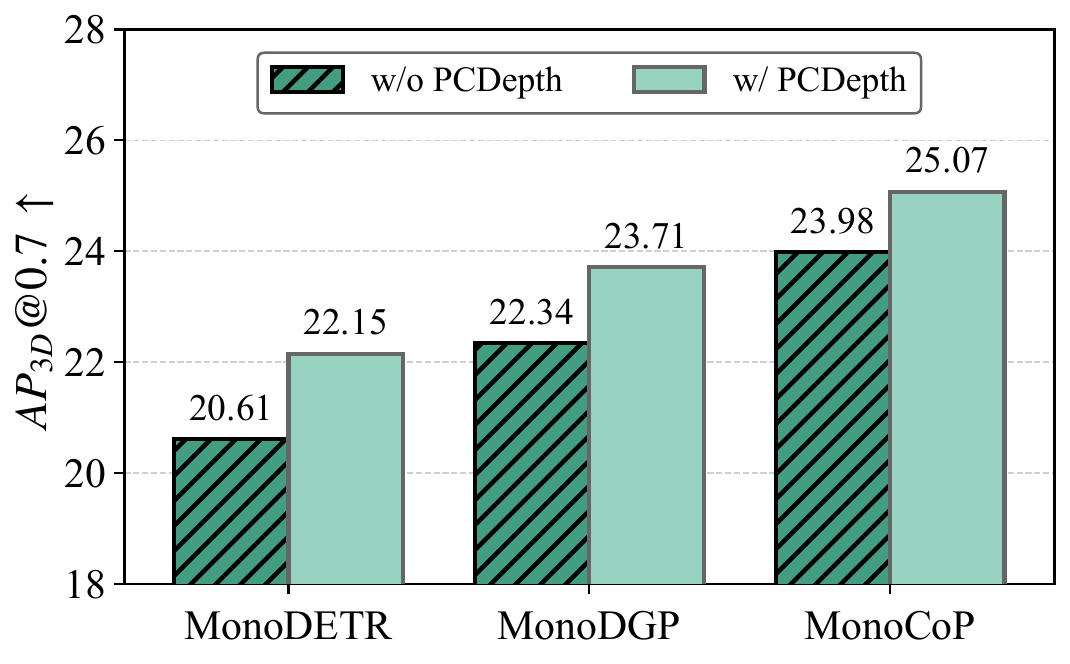}
    \vspace{-6mm}
    \small
    \caption{
        \newDepth generalizes across detectors.
        Replacing only the depth head with \newDepth yields consistent gains, indicating that the improvement stems from correcting the projection bias rather than architectural interactions.
    }
    \label{fig:prodepth-generalization}
    \vspace{-10pt}
\end{wrapfigure}
\noindent\textbf{Generality across detectors.}
We plug \newDepth into three structurally distinct \monoThreeD detectors, \monodetr~\cite{zhang2023monodetr}, \monodgp~\cite{pu2024monodgp}, and \monocop~\cite{zhang2025unleashing}, by replacing only the depth head and jointly training the full model under each baseline's original recipe. As shown in \cref{fig:prodepth-generalization}, \newDepth improves the three baselines by $+1.54\%$, $+1.37\%$, and $+1.09\%$, pushing each above its published result. The gains are positive across detectors with distinct depth-modeling designs, indicating that the improvement comes from correcting the projection bias itself rather than from architecture-specific interactions. The gain also shrinks monotonically as the baseline gets stronger ($20.61 \to 22.34 \to 23.98$), reflecting the diminishing headroom toward the performance ceiling. Even on top of the strongest baseline, however, \newDepth still extracts a non-trivial $+1.09\%$, showing that the front-surface bias remains an unresolved bottleneck.

\begin{table}[t]
    \centering
    \small
    \caption{
        Ablations on \kitti val (\apThreeD Mod., \iouThreeD $\ge 0.7$).
        \textbf{(a)} \newDepth and \geoDistill are complementary: their joint gain ($+2.24\%$) exceeds the sum of individual gains ($+1.09\% + 0.75\% = 1.84\%$).
        \textbf{(b)} Substituting the unobservable $h'_{2D}$ degrades performance; \newDepth retains the observable $h_{2D}$ and restores projection consistency through an analytic correction.
        \textbf{(c)} \mixtureBackbone outperforms the VFM-only variant despite a smaller backbone, indicating that convolutional features and VFM features carry complementary information.
        \textbf{(d)} Cosine alignment outperforms element-wise regression losses for distillation between heterogeneous backbones.
    }
    \resizebox{\linewidth}{!}{
        \begin{tabular}{cc|c|l|c|l|c|l|c}
            \midrule
            \multicolumn{3}{c|}{\textbf{(a) Components}} & \multicolumn{2}{c|}{\textbf{(b) Depth Formulation}} & \multicolumn{2}{c|}{\textbf{(c) \mixtureBackbone}} & \multicolumn{2}{c}{\textbf{(d) Alignment Loss}} \\
            \midrule
            \newDepth & \geoDistill & Mod.$\uparrow$ & Formulation          & Mod.$\uparrow$ & Configuration              & Mod.$\uparrow$ & Loss & Mod.$\uparrow$ \\
            \midrule
            \mathDash & \mathDash   & 23.98          & Direct Reg.          & 23.82          & MoB (\scaleAttention only) & 24.97          & None & 25.07 \\
            $\surd$   & \mathDash   & 25.07          & GeoDepth ($h_{2D}$)  & 24.73          & MoB (\scaleMatch only)     & 25.44          & MAE  & 25.53 \\
            \mathDash & $\surd$     & 24.73          & GeoDepth ($h'_{2D}$) & 24.18          & VFM only                   & 25.78          & MSE  & 25.69 \\
            \rowcolor{hdColor}
            $\surd$   & $\surd$     & \textbf{26.22} & \textbf{\newDepth}   & \textbf{26.22} & \textbf{MoB (full)}        & \textbf{26.22} & \textbf{\featureAlignment} & \textbf{26.22} \\
            \midrule
        \end{tabular}}
    \label{tab:ablation-all}
    \vspace{-2mm}
\end{table}

\section{Conclusion}
\label{sec:conclusion}
We identify a systematic, projection-induced bias in the conventional geometric depth formulation: substituting the \twoD bounding-box height $h_{2D}$ recovers the depth of the visible front surface rather than the object center. We address this with \methodName, which corrects the bias analytically via \newDepth and supplies the attribute fidelity it requires through \geoDistill, a teacher-student distillation strategy that transfers VFM-enriched geometric representations into a lightweight student. \methodName achieves \sota performance on \kitti, \nuscenes, \waymo, and \omniThreeD without additional inference cost. A consequence of this formulation is that depth is no longer regressed in isolation but coupled in closed form to predicted size and orientation, making attribute accuracy a direct factor in depth quality. We hope this work motivates further study of projection-consistent depth modeling in \monoThreeD.

\clearpage

{
    \small
    \bibliographystyle{plain}
    \bibliography{main}
}

\clearpage
\section*{\Large{Appendix}}
\setcounter{section}{0}
\setcounter{figure}{0}
\setcounter{table}{0}
\addcontentsline{toc}{section}{Appendix Index}
\crefalias{section}{appendix}
\makeatletter
\renewcommand{\thesection}{\Alph{section}}
\renewcommand{\theHsection}{\Alph{section}}
\renewcommand{\thefigure}{A\arabic{figure}}
\renewcommand{\theHfigure}{A\arabic{figure}}
\renewcommand{\thetable}{A\arabic{table}}
\renewcommand{\theHtable}{A\arabic{table}}
\makeatother
\renewcommand{\thetable}{A\arabic{table}}
\setcounter{equation}{0}
\renewcommand{\theequation}{A\arabic{equation}}
\section{Architecture and Implementation Details of \methodName}
\label{appendix-imple}

\subsection{Model Details}

\begin{figure}[t]
    \centering
    \includegraphics[width=\linewidth]{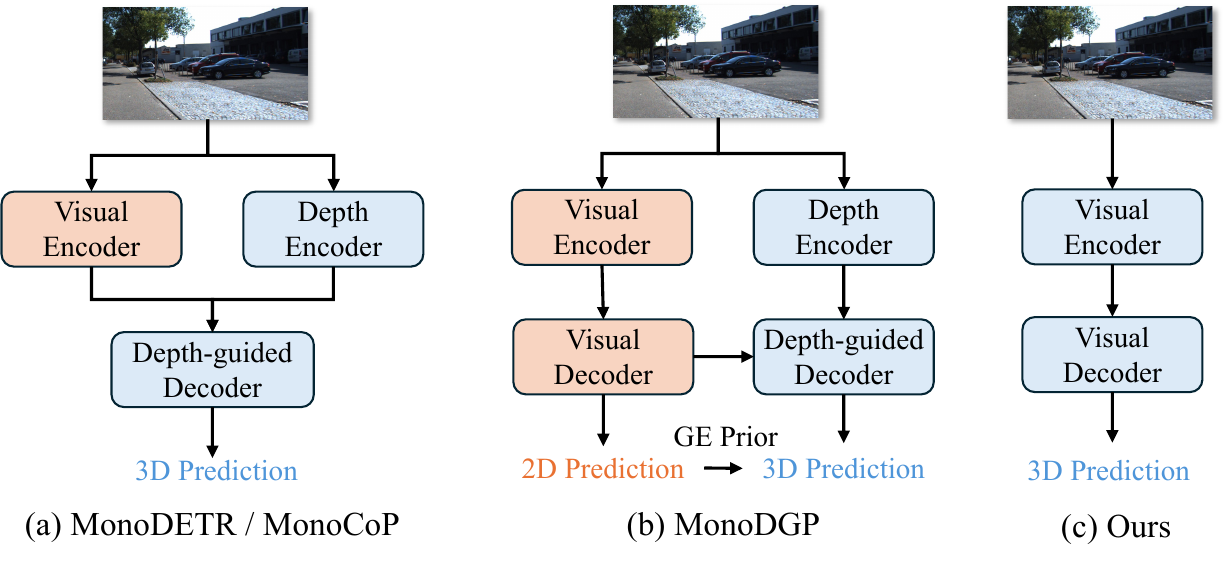}
    \small
    \caption{
        Architectural comparison between \methodName and previous methods.
        We highlight the key architectural differences between \methodName and representative monocular 3D detectors, including \monocop, \monodgp, and \monodetr.
    }
    \label{fig:arch-diff}
\end{figure}

We provide a detailed description of the architecture of \methodName.
Built upon \monocop~\cite{zhang2025unleashing}, \methodName adopts a more compact design, as illustrated in~\cref{fig:arch-diff}.
Compared with previous methods, \methodName simplifies the overall framework by jointly encoding and fusing depth and visual features.
As a result, the model only requires a standard visual encoder and decoder, leading to a cleaner and more efficient architecture.

Specifically, it consists of an image backbone for extracting visual features,
a depth predictor for generating the depth feature map,
a visual encoder that processes the fused image and depth features,
and a visual decoder that further refines the encoded representations.
On top of these components, several prediction heads are employed to estimate the \threeD object attributes,
while the proposed \newDepth formulation is used to produce the final depth prediction.

\noindent \textbf{Image Backbone.}
Our framework employs two types of image backbones.
The teacher model adopts the proposed \mixtureBackbone, whereas the student model uses a standard ResNet backbone.
Although their internal architectures differ, both backbones share the same input–output interface.
Given an input image $I \in \mathbb{R}^{H \times W \times 3}$, the backbone extracts multi-scale visual representations:
\begin{equation}
\{F_1, F_2, F_3, F_4\} = \mathcal{B}(I),
\end{equation}
where $\mathcal{B}$ denotes the backbone network and
$F_1, F_2, F_3, F_4$ are feature maps with spatial resolutions of
$\frac{1}{8}$, $\frac{1}{16}$, $\frac{1}{32}$, and $\frac{1}{64}$ of the input image, respectively, with $F_4$ obtained by a stride-2 convolution on $F_3$.
The detector consumes only $F_1, F_2, F_3$; $F_4$ is used solely by the feature alignment loss during training.
\mixtureBackbone itself follows the broader idea of adaptively combining heterogeneous experts, which has also proven effective for human recognition~\cite{zhu2025quality, zhu2026fusionagent} and monocular depth estimation~\cite{zhu2026depthagent}.

\noindent \textbf{Depth Predictor.}
To obtain depth-aware representations, following~\cite{zhang2023monodetr}, we employ a lightweight depth predictor that estimates depth features from the backbone outputs.
Given the multi-scale feature maps $\{F_1, F_2, F_3\}$ produced by the backbone, we first resize them to a unified spatial resolution of $\frac{1}{16}$ of the input image via bilinear interpolation.
The resized features are then fused through element-wise addition to aggregate multi-scale information.
Formally, the fused feature is computed as
\begin{equation}
F_D = \phi \left( \sum_{i=1}^{3} \mathcal{R}(F_i) \right),
\end{equation}
where $\mathcal{R}(\cdot)$ denotes the resizing operation to the $\frac{1}{16}$ scale and $\phi(\cdot)$ represents two stacked $3\times3$ convolution layers.
The resulting depth feature map is
\[
F_D \in \mathbb{R}^{\frac{H}{16} \times \frac{W}{16} \times C}.
\]

\noindent \textbf{Visual Encoder.}
The visual encoder aims to jointly encode and fuse the visual features from the image backbone and the depth predictor.
Specifically, it takes as input the three multi-scale backbone features $\{F_1, F_2, F_3\}$ together with the depth feature $F_D$, forming a four-level feature set.
These features are then processed by a stack of three deformable attention layers to capture cross-scale interactions and integrate depth-aware cues into the visual representations.
Formally, the encoded features are obtained as
\begin{equation}
\{E_1, E_2, E_3, E_4\} = \mathcal{E}\big(\{F_1, F_2, F_3, F_D\}\big),
\end{equation}
where $\mathcal{E}(\cdot)$ denotes the visual encoder composed of three deformable attention layers.

\noindent \textbf{Visual Decoder.}
The visual decoder refines the encoded visual representations and prepares them for the final 3D attribute prediction.
It takes the encoded multi-level features $\{E_1, E_2, E_3, E_4\}$ from the visual encoder as input and processes them through a transformer decoder consisting of one self-attention layer followed by one cross-attention layer.
The self-attention layer models interactions among object queries, while the cross-attention layer aggregates information from the encoded visual features.
Formally, the decoded representations are obtained as
\begin{equation}
Z = \mathcal{D}(Q, \{E_1, E_2, E_3, E_4\}),
\end{equation}
where $Q$ denotes the learnable object queries and $\mathcal{D}(\cdot)$ represents the visual decoder.

\noindent \textbf{Prediction Heads.}
The decoded representations $Z$ are fed into several prediction heads to estimate the 3D object attributes.
Specifically, each object query predicts a set of attributes including the 2D bounding box, object size, orientation, and corresponding confidence.
Each attribute is predicted by a lightweight multi-layer perceptron (MLP) head.
Formally, the predicted attributes are obtained as
\begin{equation}
\hat{y} = \mathcal{H}(Z),
\end{equation}
where $\mathcal{H}(\cdot)$ denotes the collection of prediction heads.

\noindent \textbf{Final Depth Estimation with \newDepth.}
Based on the predicted geometric attributes, the final object depth is computed using the proposed \newDepth formulation.
Unlike conventional geometric depth estimation that directly relies on the projected 2D height, \newDepth incorporates object size and orientation information to obtain a projection-consistent depth estimation.
Formally, the final depth is computed as
\begin{equation}
\hat{z} = \text{\newDepth}(\hat{s}, \hat{o}, \hat{h}),
\end{equation}
where $\hat{s}$, $\hat{o}$, and $\hat{h}$ denote the predicted object size, orientation, and 2D height, respectively.

\subsection{Training Objective}

Following DETR-style detectors, we adopt a bipartite matching strategy to assign predictions to ground-truth objects and supervise the matched pairs with multiple task-specific losses.
Given the predicted object set and ground-truth annotations, we first compute an optimal matching using the Hungarian algorithm, and then apply supervision on each matched pair.

\noindent \textbf{Classification Loss.}
For object classification, we employ the sigmoid focal loss to address the foreground–background imbalance:
\begin{equation}
\mathcal{L}_{cls} =
\text{FocalLoss}(p, y),
\end{equation}
where $p$ denotes the predicted class logits and $y$ represents the ground-truth class label.

\noindent \textbf{Bounding Box Loss.}
For 2D box regression, we use an $\ell_1$ loss and the GIoU loss:
\begin{equation}
\mathcal{L}_{bbox} = \| b - b^* \|_1,
\qquad
\mathcal{L}_{giou} = 1 - \text{GIoU}(b, b^*),
\end{equation}
where $b$ and $b^*$ denote the predicted and ground-truth boxes, respectively.

\noindent \textbf{3D Center Loss.}
The projected 3D object center is supervised using an $\ell_1$ loss:
\begin{equation}
\mathcal{L}_{center} =
\| c - c^* \|_1 .
\end{equation}

\noindent \textbf{Depth Loss.}
For depth prediction, we adopt an uncertainty-aware regression loss:
\begin{equation}
\mathcal{L}_{depth}
=
\sqrt{2}\, e^{-s}
|z - z^*| + s,
\end{equation}
where $z$ and $z^*$ denote the predicted and ground-truth depths, and $s$ represents the predicted log-variance that models depth uncertainty.

\noindent \textbf{Dimension Loss.}
Object size prediction is supervised using a scale-normalized $\ell_1$ loss:
\begin{equation}
\mathcal{L}_{dim}
=
\left\|
\frac{d - d^*}{d^*}
\right\|_1,
\end{equation}
where $d$ and $d^*$ denote the predicted and ground-truth object dimensions.

\noindent \textbf{Orientation Loss.}
For object orientation, we follow a classification–regression formulation.
The orientation angle is first discretized into bins for classification, followed by residual regression within each bin:
\begin{equation}
\mathcal{L}_{angle}
=
\mathcal{L}_{cls}^{angle}
+
\mathcal{L}_{reg}^{angle}.
\end{equation}

\noindent \textbf{Auxiliary Depth Map Loss.}
To further enhance depth understanding, we follow \monodetr~\cite{zhang2023monodetr} and supervise the predicted depth map with its depth-aware loss $\mathcal{L}_{depthmap}$.

\noindent \textbf{Feature Alignment Loss.}
To transfer the enriched representations from the teacher network, we employ the proposed hierarchical feature alignment loss $\mathcal{L}_{align}$, which enforces patch-wise cosine similarity between teacher and student features across all four pyramid levels $\{F_1, F_2, F_3, F_4\}$:
\begin{equation}
\mathcal{L}_{align}
= \sum_{s=1}^{4}
\frac{1}{N_s}
\sum_{i=1}^{N_s}
\left(
1 -
\frac{
\langle \hat{F}^{S}_{s,i}, F^{T}_{s,i} \rangle
}{
\|\hat{F}^{S}_{s,i}\|_2 \, \|F^{T}_{s,i}\|_2
}
\right),
\end{equation}
where $F^{T}_{s}$ and $F^{S}_{s}$ denote the teacher and student features at scale $s$, and $\hat{F}^{S}_{s}$ is the projected student feature.
This loss encourages the student network to mimic the multi-scale representations produced by the teacher.
The alignment loss is applied only during training, and the teacher network is discarded during inference.

\noindent \textbf{Overall Loss.}
The final training objective is defined as
\begin{equation}
\begin{aligned}
\mathcal{L} =
&\ \lambda_{cls}\mathcal{L}_{cls}
+ \lambda_{bbox}\mathcal{L}_{bbox}
+ \lambda_{giou}\mathcal{L}_{giou}
+ \lambda_{center}\mathcal{L}_{center} \\
&+ \lambda_{depth}\mathcal{L}_{depth}
+ \lambda_{dim}\mathcal{L}_{dim}
+ \lambda_{angle}\mathcal{L}_{angle} \\
&+ \lambda_{depthmap}\mathcal{L}_{depthmap}
+ \lambda_{align}\mathcal{L}_{align}.
\end{aligned}
\end{equation}

The loss weights are set as
$\lambda_{cls}=2$,
$\lambda_{bbox}=5$,
$\lambda_{giou}=2$,
$\lambda_{center}=10$,
$\lambda_{dim}=1$,
$\lambda_{angle}=1$,
$\lambda_{depth}=1$,
$\lambda_{depthmap}=1$,
and $\lambda_{align}=5$.

\subsection{Implementation Details}
We train the network using the AdamW optimizer with an initial learning rate of $2\times10^{-4}$ and a weight decay of $1\times10^{-4}$.
The teacher backbone is optimized with a smaller learning rate of $2\times10^{-5}$, while the student network uses a learning rate of $2\times10^{-4}$.
The learning rate follows a step decay schedule with a decay factor of $0.5$, applied at epochs 85, 125, 165, and 205, with a total of 250 training epochs.
All experiments are conducted on a single node with eight NVIDIA A100 GPUs (80GB) and a batch size of 16. A full training run on \kitti takes approximately 12 hours, while training on the larger \waymo and \omnidataset benchmarks takes one to two days.

\begin{table*}[t]
    \centering
    \caption{Detailed \kitti Leaderboard (\test) results at IoU$_{3D} \ge 0.7$. [Key: \textbf{Best}, \textcolor{myred}{\textbf{T}}: \mixtureBackbone teacher]}

    \begin{tabular}{l|c|c|ccc|ccc}
        \midrule
        \multirow{2}{*}{Method} & \multirow{2}{*} {\makecell{Extra \\ Data}} & \multirow{2}{*}{Venue} & \multicolumn{3}{c|}{ \textbf{\test}, \apThreeD (\uparrowRHDSmall) } & \multicolumn{3}{c}{ \textbf{\test}, \apBev (\uparrowRHDSmall) } \\
                                                  &       &            & Easy  & Mod.  & Hard  & Easy      & Mod.      & Hard \\
        \midrule
        \occupancymThreeD~\cite{peng2024learning} & LiDAR & CVPR~24    & 25.55 & 17.02 & 14.79 & 35.38     & 24.18     & 21.37 \\
        \opaThreeD~\cite{su2023opa}               & Depth & ICRA~23    & 24.68 & 17.17 & 14.14 & 32.50     & 23.14     & 20.30 \\
        \monotakd~\cite{liu2025monotakd}*         & LiDAR & CVPR 25    & 27.91 & 19.43 & 16.51 & 38.75     & 27.76     & 24.14 \\
        \midrule
        \monoflex~\cite{zhang2021objects}         & \multirow{14}{*}{None} & CVPR~21    & 19.94 & 13.89 & 12.07 & 28.23     & 19.75     & 16.89 \\
        \monorcnn~\cite{shi2021geometry}          &       & ICCV~21    & 18.36 & 12.65 & 10.03 & 25.48     & 18.11     & 14.10 \\
        \gupNet~\cite{lu2021geometry}             &       & CVPR~21    & 20.11 & 14.20 & 11.77 & \mathDash & \mathDash & \mathDash \\
        \deviant~\cite{kumar2022deviant}          &       & ECCV 22    & 21.88 & 14.46 & 11.89 & 29.65     & 20.44     & 17.43 \\
        \monocon{}~\cite{liu2022monocon}          &       & AAAI 22    & 22.50 & 16.46 & 13.95 & 31.12     & 22.10     & 19.00 \\
        \monodetr~\cite{zhang2023monodetr}        &       & ICCV 23    & 25.00 & 16.47 & 13.58 & 33.60     & 22.11     & 18.60 \\
        \monouni~\cite{jia2023monouni}            &       & NeurIPS 23 & 24.75 & 16.73 & 13.49 & \mathDash & \mathDash & \mathDash \\
        \fdThreeD~\cite{wu2024fd3d}               &       & AAAI~24    & 25.38 & 17.12 & 14.50 & 34.20     & 23.72     & 20.76 \\
        \monomae~\cite{jiang2024monomae}          &       & NeurIPS 24 & 25.60 & 18.84 & 16.78 & 34.15     & 24.93     & 21.76 \\
        \monocd~\cite{yan2024monocd}              &       & CVPR~24    & 25.53 & 16.59 & 14.53 & 33.41     & 22.81     & 19.57 \\
        \monodgp~\cite{pu2024monodgp}             &       & CVPR 25    & 26.35 & 18.72 & 15.97 & 35.24     & 25.23     & 22.02 \\
        \monocop~\cite{zhang2025unleashing}       &       & CVPR 26    & 27.54 & 19.11 & 16.33 & 36.77     & 25.57     & 22.62 \\
        \textbf{\methodName}                      &       & \mathDash  & \textbf{30.34} & \textbf{20.55} & \textbf{17.84} & \textbf{38.65} & \textbf{26.63} & \textbf{24.43} \\
        \midrule
        \textbf{\methodName-\textcolor{myred}{T}} & None  & \mathDash  & \textcolor{myred}{\textbf{31.78}} & \textcolor{myred}{\textbf{21.97}} & \textcolor{myred}{\textbf{19.24}} & \textcolor{myred}{\textbf{40.15}} & \textcolor{myred}{\textbf{27.64}} & \textcolor{myred}{\textbf{25.54}} \\
        \midrule
    \end{tabular}

    \label{tab:kitti-test}
\end{table*}

\section{Detailed \kitti Results}
\label{appendix-kitti}

\noindent \textbf{Leaderboard Results.}
We provide detailed results on the \kitti test benchmark.
In \kitti, the car category is the dominant class and the primary evaluation target.
Therefore, we report the detailed leaderboard results for car detection, which is also the most challenging setting in \monoThreeD.
As shown in \cref{tab:kitti-test}, \methodName outperforms previous methods by a clear margin.
For example, in the Easy setting, it improves over \monocop by $2.80\%$.
\methodName also achieves the best performance on both the Moderate and Hard levels, demonstrating consistent improvements under increasing occlusion and object difficulty.
Moreover, compared with methods that utilize additional data such as LiDAR or depth maps, \methodName still achieves higher performance on all \apThreeD metrics.

\noindent \textbf{\kitti \val Results.}
We further report detailed results on the \kitti \val set in \cref{tab:kitti-val}.
Among methods that do not rely on extra data, \methodName achieves the best performance across all evaluation metrics.
Specifically, \methodName attains $35.86\%$, $26.22\%$, and $23.58\%$ AP$_{3D}$ on the Easy, Moderate, and Hard splits, outperforming the previous best method \monocop by $3.80\%$, $2.24\%$, and $2.94\%$, respectively.
Consistent improvements are also observed for BEV detection, where \methodName achieves $46.29\%$, $33.89\%$, and $30.05\%$ AP$_{BEV}$ on the Easy, Moderate, and Hard settings.
Furthermore, the teacher model \methodName-T achieves even stronger performance, reaching $37.39\%$, $28.44\%$, and $25.02\%$ AP$_{3D}$, demonstrating the effectiveness of \mixtureBackbone.

\begin{table*}[t]
    \centering
    \caption{Detailed \kitti \val results at IoU$_{3D} \ge 0.7$.[Key: \textbf{Best}, \textcolor{myred}{\textbf{T}}: \mixtureBackbone teacher]}
    \begin{tabular}{l|c|c|ccc|ccc}
        \midrule
        \multirow{2}{*}{Method} & \multirow{2}{*} {\makecell{Extra \\ Data}} & \multirow{2}{*}{Venue} & \multicolumn{3}{c|}{ \textbf{\val}, \apThreeD (\uparrowRHDSmall) } & \multicolumn{3}{c}{ \textbf{\val}, \apBev (\uparrowRHDSmall) } \\
                                                  &       &            & Easy      & Mod.      & Hard      & Easy      & Mod.      & Hard \\
        \midrule
        \occupancymThreeD~\cite{peng2024learning} & LiDAR & CVPR~24    & 26.87     & 19.96     & 17.15     & 35.72     & 26.60     & 23.68 \\
        \opaThreeD~\cite{su2023opa}               & Depth & ICRA~23    & 24.97     & 19.40     & 16.59     & 33.80     & 25.51     & 22.13 \\
        \monotakd~\cite{liu2025monotakd}*         & LiDAR & CVPR 25    & 34.36     & 22.61     & 19.88     & 42.86     & 29.41     & 26.47 \\
        \midrule
        \monoflex~\cite{zhang2021objects}         & \multirow{13}{*}{None} & CVPR~21    & 23.64     & 17.51     & 14.83     & \mathDash & \mathDash & \mathDash \\
        \monorcnn~\cite{shi2021geometry}          &       & ICCV~21    & \mathDash & \mathDash & \mathDash & \mathDash & \mathDash & \mathDash \\
        \gupNet~\cite{lu2021geometry}             &       & CVPR~21    & 22.76     & 16.46     & 13.72     & 31.07     & 22.94     & 19.75 \\
        \deviant~\cite{kumar2022deviant}          &       & ECCV 22    & 24.63     & 16.54     & 14.52     & 32.60     & 23.04     & 19.99 \\
        \monocon{}~\cite{liu2022monocon}          &       & AAAI 22    & 26.33     & 19.01     & 15.98     & \mathDash & \mathDash & \mathDash \\
        \monodetr~\cite{zhang2023monodetr}        &       & ICCV 23    & 28.84     & 20.61     & 16.38     & 37.86     & 26.95     & 22.80 \\
        \monouni~\cite{jia2023monouni}            &       & NeurIPS 23 & 24.51     & 17.18     & 14.01     & \mathDash & \mathDash & \mathDash \\
        \fdThreeD~\cite{wu2024fd3d}               &       & AAAI~24    & 28.22     & 20.23     & 17.04     & 36.98     & 26.77     & 23.16 \\
        \monomae~\cite{jiang2024monomae}          &       & NeurIPS 24 & 30.29     & 20.90     & 17.61     & 40.26     & 27.08     & 23.14 \\
        \monocd~\cite{yan2024monocd}              &       & CVPR~24    & 26.45     & 19.37     & 16.38     & 34.60     & 24.96     & 21.51 \\
        \monodgp~\cite{pu2024monodgp}             &       & CVPR 25    & 30.76     & 22.34     & 19.02     & 39.40     & 28.20     & 24.42 \\
        \monocop~\cite{zhang2025unleashing}       &       & CVPR 26    & 32.06     & 23.98     & 20.64     & 41.92     & 30.75     & 27.06 \\
        \textbf{\methodName}                      &       & \mathDash  & \textbf{35.86} & \textbf{26.22} & \textbf{23.58} & \textbf{46.29} & \textbf{33.89} & \textbf{30.05} \\
        \midrule
        \textbf{\methodName-\textcolor{myred}{T}} & None  & \mathDash  & \textcolor{myred}{\textbf{37.39}} & \textcolor{myred}{\textbf{28.44}} & \textcolor{myred}{\textbf{25.02}} & \textcolor{myred}{\textbf{47.95}} & \textcolor{myred}{\textbf{35.36}} & \textcolor{myred}{\textbf{31.50}} \\
        \midrule
    \end{tabular}

    \label{tab:kitti-val}
\end{table*}

\section{Detailed \nuscenes Results}
\label{appendix-nuscenes}

On the \nuscenes \val set, \methodName achieves new \sota performance under both IoU thresholds $0.7$ and $0.5$ across \threeD and \bev detection.
Under the primary setting IoU $\ge 0.7$, \methodName improves \apThreeD Mod. by $+2.32\%$ over \monocop, with consistent gains in BEV detection.
At IoU $\ge 0.5$, the advantage becomes more pronounced, demonstrating strong robustness under relaxed localization constraints. The teacher \methodName-T yields further gains in \apThreeD under both IoU thresholds, and the distilled student recovers about half of the teacher's improvement over \monocop, demonstrating the effectiveness of \geoDistill.

\begin{table}[t]
    \centering
    \caption{
        \nuscenes \val results. \methodName achieves \sota performance on both \threeD and \bev detection under IoU thresholds 0.7 and 0.5.
        [Key: \textbf{Best}, \underline{Second Best}, \textcolor{myblue}{\textbf{Gain}}, \textcolor{myred}{\textbf{T}}: \mixtureBackbone teacher]
    }
    \tabcolsep=0.25cm
    \resizebox{0.9\textwidth}{!}{
        \begin{tabular}{l|cc|cc|cc|cc}
            \midrule
            \multirow{3}{*}{Method} & \multicolumn{4}{c|}{\textbf{\iouThreeD $\ge$ 0.7}} & \multicolumn{4}{c}{\textbf{\iouThreeD $\ge$ 0.5}} \\
            \cmidrule(lr){2-5} \cmidrule(lr){6-9}
            & \multicolumn{2}{c|}{\apThreeD $\uparrow$} & \multicolumn{2}{c|}{\apBev $\uparrow$} & \multicolumn{2}{c|}{\apThreeD $\uparrow$} & \multicolumn{2}{c}{\apBev $\uparrow$} \\
            \cmidrule(lr){2-3} \cmidrule(lr){4-5} \cmidrule(lr){6-7} \cmidrule(lr){8-9}
                                                      & Easy  & Mod. & Easy  & Mod.  & Easy  & Mod.  & Easy  & Mod. \\
            \midrule
            \gupNet~\cite{lu2021geometry}             & 8.50  & 7.40 & 14.21 & 12.81 & 29.03 & 26.16 & 33.42 & 30.23 \\
            \deviant~\cite{kumar2022deviant}          & 9.69  & 8.33 & 16.28 & 14.36 & 31.47 & 28.22 & 35.61 & 31.93 \\
            \monodetr~\cite{zhang2023monodetr}        & 9.53  & 8.19 & 16.39 & 14.41 & 31.81 & 28.35 & 35.70 & 31.96 \\
            \monodgp~\cite{pu2024monodgp}             & 10.04 & 8.78 & 16.55 & 14.53 & 29.56 & 26.17 & 32.67 & 29.44 \\
            \monocop~\cite{zhang2025unleashing}       & \underline{10.85} & \underline{9.71} & \underline{17.83} & \underline{15.86} & \underline{33.70} & \underline{29.91} & \underline{37.44} & \underline{34.01} \\
            \textbf{\methodName}                      & \textbf{13.75} & \textbf{12.03} & \textbf{20.97} & \textbf{18.49} & \textbf{38.50} & \textbf{34.11} & \textbf{42.03} & \textbf{38.49} \\
            \rowcolor{hdColor}
            \textcolor{myblue}{\textbf{$\Delta$}}     & \textcolor{myblue}{\textbf{+2.90}} & \textcolor{myblue}{\textbf{+2.32}} & \textcolor{myblue}{\textbf{+3.14}} & \textcolor{myblue}{\textbf{+2.63}} & \textcolor{myblue}{\textbf{+4.80}} & \textcolor{myblue}{\textbf{+4.20}} & \textcolor{myblue}{\textbf{+4.59}} & \textcolor{myblue}{\textbf{+4.48}} \\
            \midrule
            \textbf{\methodName-\textcolor{myred}{T}} & \textcolor{myred}{\textbf{16.52}} & \textcolor{myred}{\textbf{14.57}} & \textcolor{myred}{\textbf{25.06}} & \textcolor{myred}{\textbf{22.40}} & \textcolor{myred}{\textbf{41.81}} & \textcolor{myred}{\textbf{37.22}} & \textcolor{myred}{\textbf{45.10}} & \textcolor{myred}{\textbf{41.41}} \\
            \midrule
        \end{tabular}
    }

    \label{exp:nuscenes-val}
\end{table}

\begin{table*}[t]
    \centering
    \caption{\waymo \valOne Vehicle results with \iouThreeD $\geq 0.5$. [Key: \textbf{Best}, \textcolor{myred}{\textbf{T}}: \mixtureBackbone teacher]}
    \setlength\tabcolsep{3pt}
    \resizebox{1\textwidth}{!}{
        \begin{tabular}{l|c|cccc|cccc}
            \midrule
            \multirow{2}{*}{Method} & \multirow{2}{*}{Difficulty} & \multicolumn{4}{c|}{\aphThreeD (\%)} & \multicolumn{4}{c}{\apThreeD (\%)} \\
                                                                     &  & All   & 0-30  & 30-50 & 50-$\infty$ & All   & 0-30  & 30-50 & 50-$\infty$ \\
            \midrule
            \gupNet~\cite{lu2021geometry} in \cite{kumar2022deviant} & \multirow{8}{*}{Level 1} & 9.94  & 24.59 & 4.78  & 0.22        & 10.02 & 24.78 & 4.84  & 0.22 \\
            \deviant~\cite{kumar2022deviant}                         &  & 10.89 & 26.64 & 5.08  & 0.18        & 10.98 & 26.85 & 5.13  & 0.18 \\
            \monodetr~\cite{zhang2023monodetr}                       &  & 9.60  & 23.58 & 4.67  & 0.99        & 9.68  & 23.78 & 4.72  & 1.00 \\
            \monouni~\cite{jia2023monouni}                           &  & 10.73 & 26.30 & 3.98  & 0.55        & 10.98 & 26.63 & 4.04  & 0.57 \\
            \monodgp~\cite{pu2024monodgp}                            &  & 9.84  & 23.73 & 5.01  & 0.98        & 10.06 & 24.01 & 5.06  & 0.99 \\
            \monocop~\cite{zhang2025unleashing}                      &  & 11.65 & 27.35 & 5.97  & 1.46        & 11.76 & 27.59 & 6.03  & 1.48 \\
            \textbf{\methodName}                                     &  & 12.05 & 31.99 & 7.59  & 1.66        & 12.17 & 32.36 & 7.64  & 1.68 \\
            \textbf{\methodName-\textcolor{myred}{T}}                &  & \textcolor{myred}{\textbf{15.23}} & \textcolor{myred}{\textbf{36.52}} & \textcolor{myred}{\textbf{8.50}} & \textcolor{myred}{\textbf{1.90}} & \textcolor{myred}{\textbf{15.34}} & \textcolor{myred}{\textbf{36.79}} & \textcolor{myred}{\textbf{8.57}} & \textcolor{myred}{\textbf{1.92}} \\
            \midrule
            \gupNet~\cite{lu2021geometry} in \cite{kumar2022deviant} & \multirow{8}{*}{Level 2} & 9.31  & 24.50 & 4.62  & 0.19        & 9.39  & 24.69 & 4.67  & 0.19 \\
            \deviant~\cite{kumar2022deviant}                         &  & 10.20 & 26.54 & 4.90  & 0.16        & 10.29 & 26.75 & 4.95  & 0.16 \\
            \monodetr~\cite{zhang2023monodetr}                       &  & 9.00  & 23.49 & 4.51  & 0.86        & 9.08  & 23.70 & 4.55  & 0.87 \\
            \monouni~\cite{jia2023monouni}                           &  & 10.24 & 26.24 & 3.89  & 0.51        & 10.38 & 26.57 & 3.95  & 0.53 \\
            \monodgp~\cite{pu2024monodgp}                            &  & 9.32  & 23.65 & 4.84  & 0.85        & 9.43  & 23.92 & 4.88  & 0.86 \\
            \monocop~\cite{zhang2025unleashing}                      &  & 10.93 & 27.25 & 5.76  & 1.27        & 11.03 & 27.49 & 5.82  & 1.29 \\
            \textbf{\methodName}                                     &  & 11.49 & 31.91 & 7.43  & 1.42        & 11.60 & 32.27 & 7.48  & 1.44 \\
            \textbf{\methodName-\textcolor{myred}{T}}                &  & \textcolor{myred}{\textbf{14.42}} & \textcolor{myred}{\textbf{36.39}} & \textcolor{myred}{\textbf{8.25}} & \textcolor{myred}{\textbf{1.67}} & \textcolor{myred}{\textbf{14.53}} & \textcolor{myred}{\textbf{36.66}} & \textcolor{myred}{\textbf{8.31}} & \textcolor{myred}{\textbf{1.69}} \\
            \midrule
        \end{tabular}}
    \label{tab:waymo_iou05}
\end{table*}

\begin{table*}[h]
    \centering
    \caption{\waymo \valOne Vehicle results with \iouThreeD $\geq 0.7$. [Key: \textbf{Best}, \textcolor{myred}{\textbf{T}}: \mixtureBackbone teacher]}
    \setlength\tabcolsep{3pt}
    \resizebox{1\textwidth}{!}{
        \begin{tabular}{l|c|cccc|cccc}
            \midrule
            \multirow{2}{*}{Method} & \multirow{2}{*}{Difficulty} & \multicolumn{4}{c|}{\aphThreeD (\%)} & \multicolumn{4}{c}{\apThreeD (\%)} \\
                                                                     &  & All  & 0-30 & 30-50 & 50-$\infty$ & All  & 0-30 & 30-50 & 50-$\infty$ \\
            \midrule
            \gupNet~\cite{lu2021geometry} in \cite{kumar2022deviant} & \multirow{8}{*}{Level 1} & 2.27 & 6.11 & 0.80  & 0.03        & 2.28 & 6.15 & 0.81  & 0.03 \\
            \deviant~\cite{kumar2022deviant}                         &  & 2.67 & 6.90 & 0.98  & 0.02        & 2.69 & 6.95 & 0.99  & 0.02 \\
            \monodetr~\cite{zhang2023monodetr}                       &  & 2.10 & 5.94 & 0.73  & 0.12        & 2.11 & 5.99 & 0.73  & 0.12 \\
            \monouni~\cite{jia2023monouni}                           &  & 3.16 & 8.50 & 0.86  & 0.12        & 3.20 & 8.61 & 0.87  & 0.13 \\
            \monodgp~\cite{pu2024monodgp}                            &  & 2.39 & 6.62 & 0.84  & 0.12        & 2.41 & 6.67 & 0.84  & 0.12 \\
            \monocop~\cite{zhang2025unleashing}                      &  & 2.70 & 7.38 & 1.06  & 0.16        & 2.72 & 7.44 & 1.07  & 0.16 \\
            \textbf{\methodName}                                     &  & 2.96 & 9.15 & 1.04  & 0.20        & 2.98 & 9.21 & 1.12  & 0.20 \\
            \textbf{\methodName-\textcolor{myred}{T}}                &  & \textcolor{myred}{\textbf{3.16}} & \textcolor{myred}{\textbf{10.30}} & \textcolor{myred}{\textbf{1.26}} & \textcolor{myred}{\textbf{0.22}} & \textcolor{myred}{\textbf{3.18}} & \textcolor{myred}{\textbf{10.36}} & \textcolor{myred}{\textbf{1.27}} & \textcolor{myred}{\textbf{0.22}} \\
            \midrule
            \gupNet~\cite{lu2021geometry} in \cite{kumar2022deviant} & \multirow{8}{*}{Level 2} & 2.12 & 6.08 & 0.77  & 0.02        & 2.14 & 6.13 & 0.78  & 0.02 \\
            \deviant~\cite{kumar2022deviant}                         &  & 2.50 & 6.87 & 0.94  & 0.02        & 2.52 & 6.93 & 0.95  & 0.02 \\
            \monodetr~\cite{zhang2023monodetr}                       &  & 1.97 & 5.92 & 0.70  & 0.10        & 1.98 & 5.96 & 0.71  & 0.10 \\
            \monouni~\cite{jia2023monouni}                           &  & 3.00 & 8.48 & 0.84  & 0.12        & 3.04 & 8.59 & 0.85  & 0.12 \\
            \monodgp~\cite{pu2024monodgp}                            &  & 2.24 & 6.59 & 0.81  & 0.10        & 2.26 & 6.65 & 0.81  & 0.10 \\
            \monocop~\cite{zhang2025unleashing}                      &  & 2.53 & 7.35 & 1.02  & 0.14        & 2.55 & 7.41 & 1.03  & 0.14 \\
            \textbf{\methodName}                                     &  & 2.84 & 9.13 & 1.00  & 0.17        & 2.85 & 9.18 & 1.11  & 0.17 \\
            \textbf{\methodName-\textcolor{myred}{T}}                &  & \textcolor{myred}{\textbf{2.96}} & \textcolor{myred}{\textbf{10.26}} & \textcolor{myred}{\textbf{1.22}} & \textcolor{myred}{\textbf{0.19}} & \textcolor{myred}{\textbf{2.98}} & \textcolor{myred}{\textbf{10.33}} & \textcolor{myred}{\textbf{1.23}} & \textcolor{myred}{\textbf{0.19}} \\
            \midrule
        \end{tabular}}
    \label{tab:waymo_iou07}
\end{table*}

\section{Detailed \waymo Results}
\label{appendix-waymo}

\cref{tab:waymo_iou05,tab:waymo_iou07} present detailed evaluation results on the \waymo benchmark under two IoU thresholds (\(0.5\) and \(0.7\)), covering both Level~1 and Level~2 settings.

\noindent\textbf{Results at IoU \( \geq 0.5 \).}
Under the looser IoU threshold, \methodName already achieves clear improvements over previous monocular methods.
For Level~1, \methodName obtains \(12.17\%\) \apThreeD and \(12.05\%\) \aphThreeD, surpassing MonoCoP (11.76\% / 11.65\%) by about \(+0.4\) points.
In relative terms, the improvement is largest in the 30--50\,m range, where \methodName reaches \(7.64\%\) \apThreeD, substantially outperforming the previous best \(6.03\%\).
Beyond 50\,m, it also achieves the strongest performance with \(1.68\%\).
With the high-capacity teacher architecture, \methodName-\textcolor{myred}{T} further pushes the performance to \(15.34\%\) \apThreeD and \(15.23\%\) \aphThreeD on Level~1,
improving over MonoCoP by more than \(+3.5\) points.
Notably, the teacher model achieves \(8.57\%\) \apThreeD at 30--50\,m and \(1.92\%\) beyond 50\,m, establishing new state-of-the-art results across all distance ranges.
A similar trend is observed for Level~2.
\methodName reaches \(11.60\%\) \apThreeD and \(11.49\%\) \aphThreeD, consistently outperforming previous monocular approaches.
The teacher model further improves the results to \(14.53\%\) \apThreeD and \(14.42\%\) \aphThreeD, again achieving the best performance across all distance splits.

\noindent\textbf{Results at IoU \( \geq 0.7 \).}
Under the stricter localization criterion, \methodName achieves the best results in the 0--30\,m and 50\,m+ ranges and the best \apThreeD in the 30--50\,m range.
For Level~1, \methodName achieves \(2.98\%\) \apThreeD and \(2.96\%\) \aphThreeD, exceeding MonoCoP (2.72\% / 2.70\%).
The teacher model further improves the results to \(3.18\%\) \apThreeD and \(3.16\%\) \aphThreeD, with the strongest performance across all distance ranges.
The same trend holds for Level~2, where \methodName achieves \(2.85\%\) \apThreeD and \(2.84\%\) \aphThreeD,
while \methodName-\textcolor{myred}{T} further improves to \(2.98\%\) \apThreeD and \(2.96\%\) \aphThreeD.

Overall, \methodName improves detection accuracy at both evaluation levels, with gains in nearly all distance ranges.
The gains are largest within 50\,m, where the front-surface offset is a dominant error source; beyond this range, depth noise that grows with distance limits the benefit of any correction.

\begin{table}[t]
    \centering
    \caption{AP$_{\text{3D}}$ performance across 5 categories on \kitti. [Key: \textbf{Best}, \textcolor{myred}{\textbf{T}}: \mixtureBackbone teacher]}
    \resizebox{0.8\textwidth}{!}{
        \begin{tabular}{l|c|c|c|c|c|c}
            \midrule
            \multirow{2}{*}{Method} & \multicolumn{6}{c}{\apThreeD on Category} \\
                                                      & Cyclist & Truck & Pedestrian & Van   & Car   & Avg \\
            \midrule
            \methodName                               & 28.85   & 27.69 & 36.29      & 21.22 & 76.35 & 38.09 \\
            \textbf{\methodName-\textcolor{myred}{T}} & \textcolor{myred}{27.68} & \textcolor{myred}{27.25} & \textcolor{myred}{38.23} & \textcolor{myred}{28.62} & \textcolor{myred}{76.18} & \textcolor{myred}{39.60} \\
            \midrule
        \end{tabular}}

    \label{tab:cate-kit}
\end{table}

\begin{table}[t]
    \centering
    \caption{AP$_{\text{3D}}$ performance across 9 categories on nuScenes. [Key: \textbf{Best}, \textcolor{myred}{\textbf{T}}: \mixtureBackbone teacher]}
    \resizebox{1\textwidth}{!}{
        \begin{tabular}{l|c|c|c|c|c|c|c|c|c|c}
            \midrule
            \multirow{2}{*}{Method} & \multicolumn{10}{c}{\apThreeD on Category} \\
                                                      & Pedestrian & Motorcycle & Truck & Trailer & Car   & Traffic\,Cone & Bus   & Barrier & Bicycle & Avg \\
            \midrule
            \textbf{\methodName}                      & 14.45      & 24.62      & 45.04 & 26.93   & 77.64 & 11.52         & 70.52 & 40.57   & 21.53   & 36.99 \\
            \textbf{\methodName-\textcolor{myred}{T}} & \textcolor{myred}{15.15} & \textcolor{myred}{27.62} & \textcolor{myred}{44.41} & \textcolor{myred}{27.39} & \textcolor{myred}{78.93} & \textcolor{myred}{17.18} & \textcolor{myred}{67.98} & \textcolor{myred}{51.04} & \textcolor{myred}{23.01} & \textcolor{myred}{39.20} \\
            \midrule
        \end{tabular}}
    \label{tab:cate-nu}
\end{table}

\section{Additional \omnidatasetout Results}
\label{appendix-omni3d}
While the Omni3D benchmark reports only the averaged \apThreeD across all categories, we additionally provide per-category results for completeness.
For KITTI, we report performance across its five evaluated categories (see~\cref{tab:cate-kit}), and for nuScenes, we include all nine object categories in the benchmark (see~\cref{tab:cate-nu}).
Per-category results of the teacher \methodName-T are also reported.

\begin{figure}[t]
    \centering
    \begin{minipage}{0.6\textwidth}
        \centering
        \includegraphics[width=\linewidth]{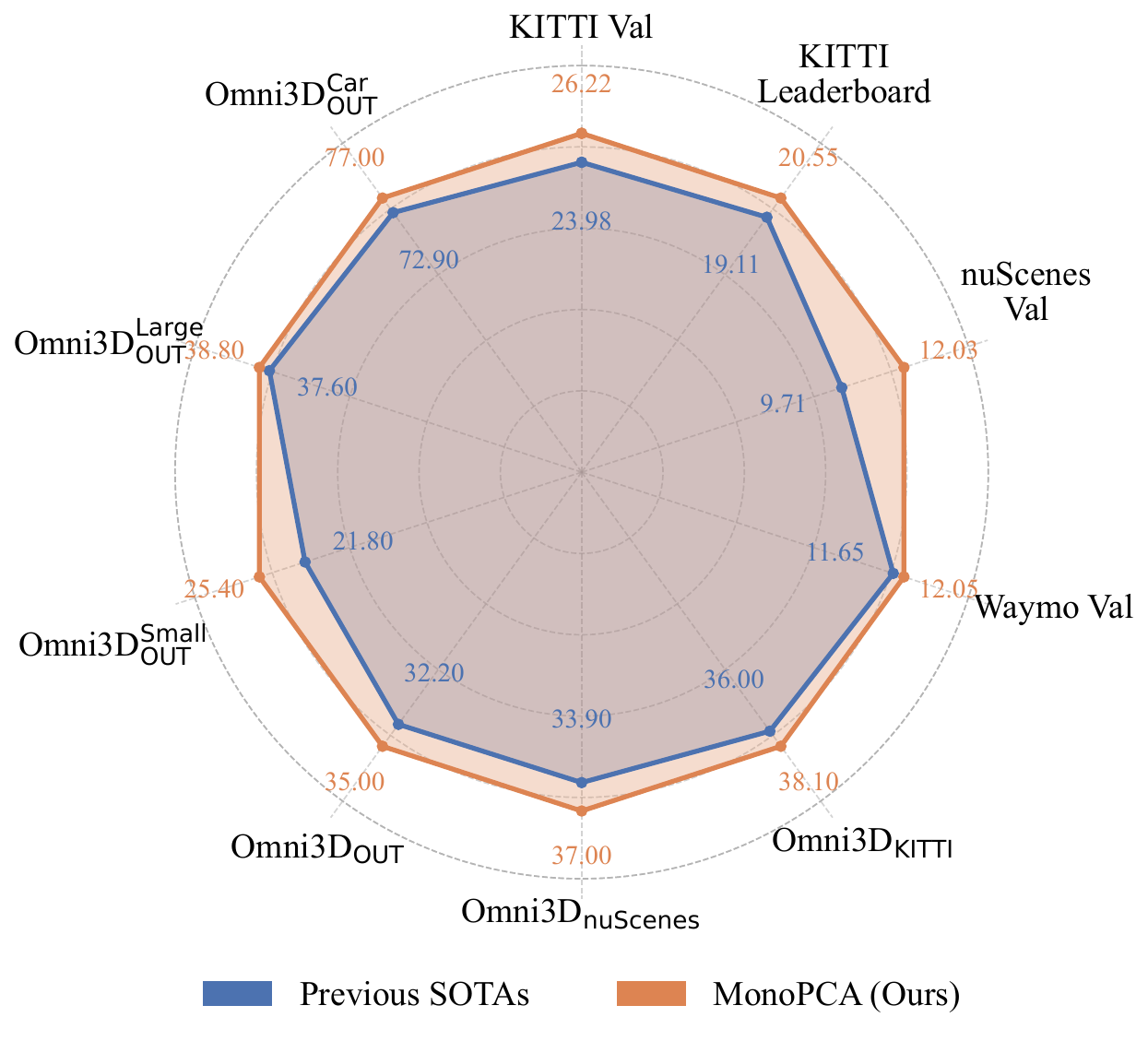}
        \caption{\methodName across ten benchmarks.}
        \label{fig:radar}
    \end{minipage}

\end{figure}

We also present a radar chart (see~\cref{fig:radar}) to illustrate the overall performance of \methodName across ten evaluation benchmarks.
As shown in the figure, \methodName consistently outperforms previous state-of-the-art methods across diverse datasets, including KITTI, nuScenes, Waymo, and Omni3D, demonstrating its strong generalization ability under different evaluation settings.

\section{Additional Visualization Results}
\label{appendix-vis}
\begin{figure*}[h]
    \centering
    \includegraphics[width=0.85\linewidth]{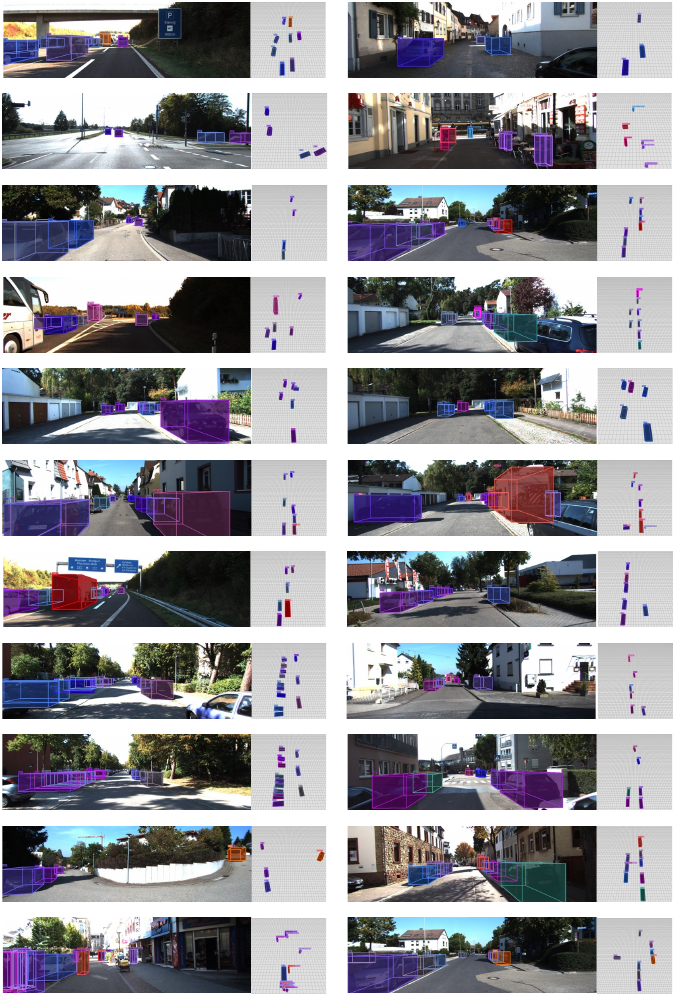}
    \caption{Qualitative results on \omnidatasetkitti. Following~\cite{brazil2023omni3d}, we show the \threeD predictions overlaid on the image (Left), and the corresponding top view with a $1$m$\times$$1$m ground grid (Right). \methodName presents more accurate results for \textbf{distant and small objects.}}
    \label{fig:qualitative-kit-1}
\end{figure*}

\begin{figure*}[h]
    \centering
    \includegraphics[width=0.85\linewidth]{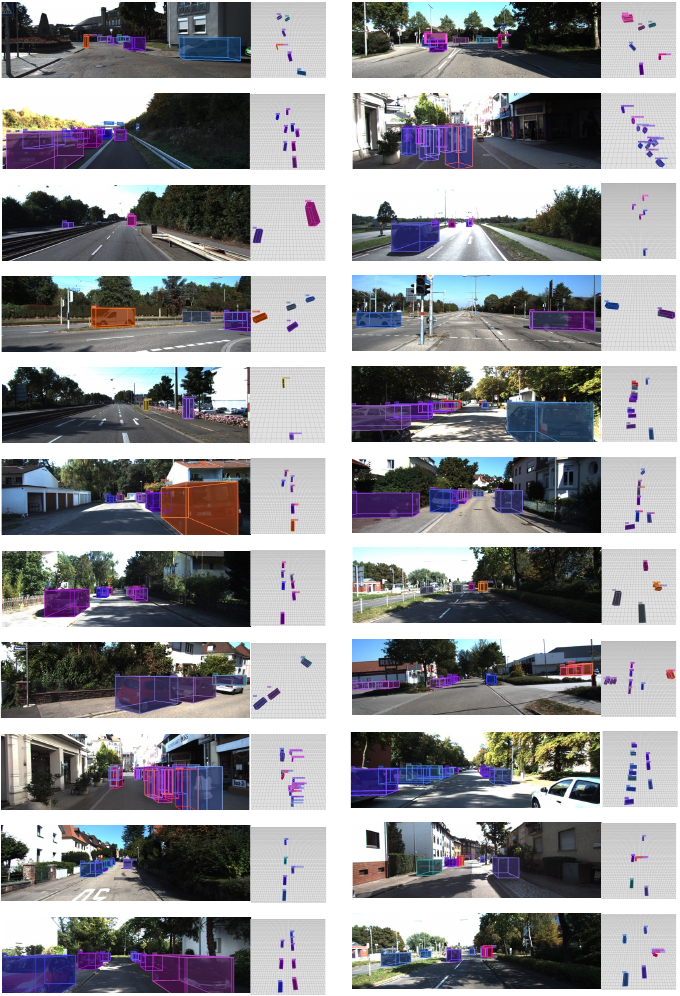}
    \caption{Qualitative results on \omnidatasetkitti. Following~\cite{brazil2023omni3d}, we show the \threeD predictions overlaid on the image (Left), and the corresponding top view with a $1$m$\times$$1$m ground grid (Right). \methodName presents more accurate results for \textbf{distant and small objects.}}
    \label{fig:qualitative-kit-2}
\end{figure*}

\begin{figure*}[h]
    \centering
    \includegraphics[width=0.7\linewidth]{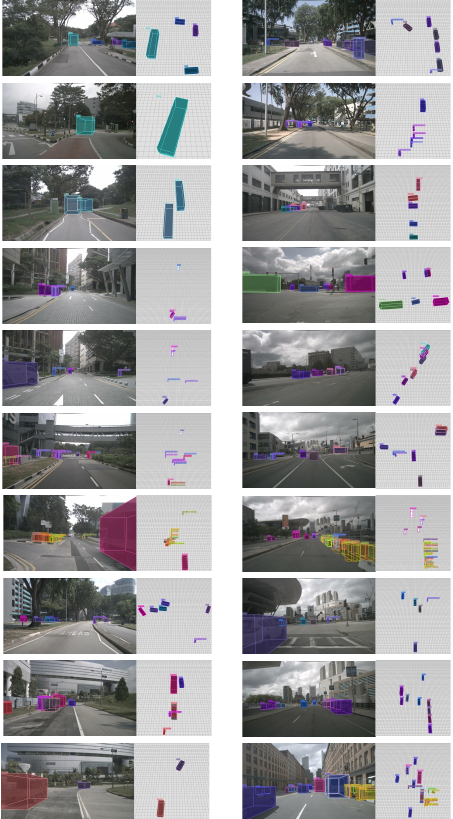}
    \caption{Qualitative results on \omnidatasetnu. Following~\cite{brazil2023omni3d}, we show the \threeD predictions overlaid on the image (Left), and the corresponding top view with a $1$m$\times$$1$m ground grid (Right). \methodName presents more accurate results for \textbf{distant and small objects.}}
    \label{fig:qualitative-nu-1}
\end{figure*}

\begin{figure*}[h]
    \centering
    \includegraphics[width=0.7\linewidth]{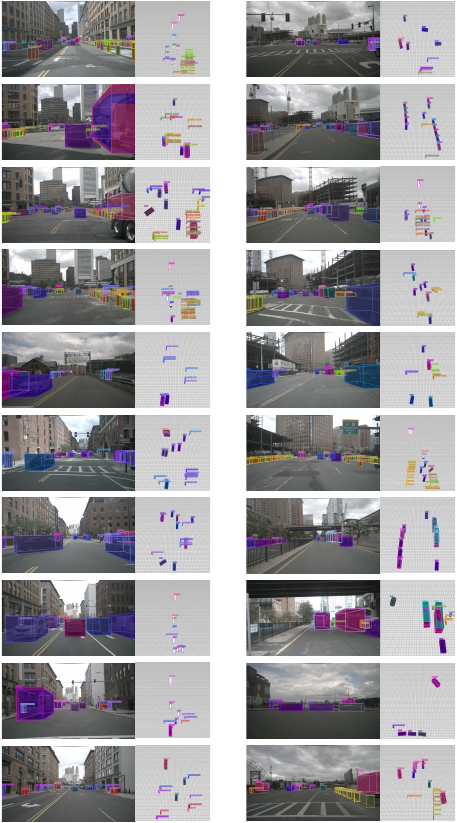}
    \caption{Qualitative results on \omnidatasetnu. Following~\cite{brazil2023omni3d}, we show the \threeD predictions overlaid on the image (Left), and the corresponding top view with a $1$m$\times$$1$m ground grid (Right). \methodName presents more accurate results for \textbf{distant and small objects.}}
    \label{fig:qualitative-nu-2}
\end{figure*}

Following~\cite{brazil2023omni3d}, we visualize the \threeD predictions overlaid on the input image (Left) and the corresponding bird’s eye view with a $1$m $\times$ $1$m ground grid (Right). \methodName produces more accurate predictions, particularly for distant and small objects.
Additional qualitative examples on the \kitti dataset are provided in \cref{fig:qualitative-kit-1} and \cref{fig:qualitative-kit-2}.
Qualitative results on the \nuscenes dataset are shown in \cref{fig:qualitative-nu-1} and \cref{fig:qualitative-nu-2}.


\section{Additional Ablation Study}
\label{appendix-ablation}

\subsection{Things We Tried That Did Not Make it into the Main Algorithm}

\noindent\textbf{From Deformable Attention to Full Attention.}
Following \monocop, we adopt deformable attention in both the visual encoder and the decoder.
Compared with vanilla Transformer attention, deformable attention is significantly more efficient, as it attends to a sparse set of sampled key locations rather than all spatial tokens.
However, this design also limits the ability to capture global context.
Motivated by this observation, and by the success of full attention in modeling long-range and cross-modal dependencies in other domains~\cite{zhang2023tile, zhang2022vrformer}, we explored replacing deformable attention with standard full attention in both the encoder and decoder to enable global feature interactions.

In practice, however, this modification did not lead to improvements.
Using full attention substantially increases the computational cost and memory consumption due to its quadratic complexity with respect to the number of tokens.
More importantly, we observed a consistent performance drop compared with the deformable attention design.
We conjecture that the sparse sampling mechanism in deformable attention provides an implicit inductive bias that helps the model focus on geometrically relevant regions, whereas full attention introduces excessive global interactions that may dilute these spatial cues.
Therefore, we retain the deformable attention architecture in the final design due to its superior efficiency and better empirical performance.

\noindent\textbf{Predicting $h'_{2D}$ for GeoDepth.}
To eliminate the projection-induced bias in the GeoDepth formulation, a straightforward idea is to directly predict the projective height $h'_{2D}$ and use it in place of the conventional \twoD bounding box height $h_{2D}$.
This modification is conceptually simple and can correct the bias in the geometric relation.

However, in practice this approach does not improve detection performance and often leads to unstable training.
We observe that $h'_{2D}$ is inherently depth-dependent and varies with object pose and occlusion, making it difficult to estimate reliably from image features alone.
As a result, direct regression of $h'_{2D}$ introduces additional prediction uncertainty, which propagates to the depth estimation and degrades the final \threeD detection accuracy.
Therefore, instead of directly predicting $h'_{2D}$, our proposed formulation models the geometric correction explicitly and in closed form through \newDepth, which achieves more stable optimization and better empirical performance.

\noindent\textbf{Integrating VFM without FPN.}
Another design we explored is to directly integrate the last-layer features from the Vision Foundation Model (VFM) without constructing a multi-scale feature pyramid.
This approach simplifies the architecture and reduces additional feature processing, as the detection head operates directly on the high-level semantic representation produced by the VFM.

However, we find that relying on a single-scale feature significantly degrades detection performance.
Object detection, especially in \monoThreeD settings, benefits from multi-scale representations to capture objects at different sizes and distances.
Using only the final VFM feature map removes this multi-scale structure and weakens the model's ability to localize objects accurately.
Therefore, in the final design we retain the FPN-based multi-scale feature construction when integrating VFM features, which provides richer spatial representations and leads to consistently better detection performance.

\noindent\textbf{Using Pixel Shuffle for Token Reduction.}
We also explored reducing the number of image tokens using a pixel-shuffle style operation.
Specifically, neighboring feature maps are first merged and then projected through a lightweight projector to reduce the channel dimension.
This operation effectively decreases the number of image tokens and thus lowers the computational cost of subsequent transformer layers.

However, we observe that this design consistently leads to a noticeable performance drop, typically around \(2\%\sim3\%\) in detection accuracy.
We conjecture that merging neighboring features in this manner removes fine-grained spatial details that are important for accurate object localization in \monoThreeD detection.
Therefore, we do not adopt this token reduction strategy.

\noindent\textbf{Replacing PositionEmbeddingSine with RoPE.}
We also explored replacing the commonly used PositionEmbeddingSine with Rotary Position Embedding (RoPE), which has been widely adopted in Transformer architectures for its ability to encode relative positional information.

However, this modification does not improve detection performance in our setting.
In practice, we observe a slight performance degradation compared with the original sine positional embedding.
We conjecture that RoPE is more suitable for sequence modeling tasks, while dense prediction tasks such as \monoThreeD detection rely more heavily on explicit spatial priors.
The PositionEmbeddingSine embedding provides a stable and well-structured spatial encoding for image features, which appears to be more effective for our detection framework.
As a result, we retain PositionEmbeddingSine in the final model design.

\subsection{More Depth Design}
\label{appendix-ablation-depth}

\cref{tab:ablation-more-depth} further compares different depth design strategies.
We evaluate several representative formulations, including keypoint-based depth prediction~\cite{shi2021geometry}, 2D height uncertainty modeling~\cite{lu2021geometry}, and bias regression~\cite{pu2024monodgp}.
As shown in the table, progressively improving the depth formulation consistently enhances detection performance.
Keypoint prediction achieves \(30.73\%\), \(23.25\%\), and \(19.85\%\) on Easy, Moderate, and Hard, while introducing 2D height uncertainty improves the results to \(32.52\%\), \(25.06\%\), and \(21.19\%\).
Regressing the bias further increases the performance to \(33.84\%\), \(25.73\%\), and \(22.30\%\).
Our \textbf{\newDepth} achieves the best results with \(35.86\%\), \(26.22\%\), and \(23.58\%\), demonstrating that a more principled depth formulation leads to more accurate depth estimation and better \threeD detection performance.

\begin{table}[t]
    \centering

    \begin{minipage}{0.48\columnwidth}
        \centering
        \small
        \caption{
            Ablation on depth design.
        }
        \setlength{\tabcolsep}{3pt}
        \resizebox{1\linewidth}{!}{
            \begin{tabular}{l|ccc}
                \midrule
                More Depth Design         & Easy$\uparrow$ & Mod.$\uparrow$ & Hard$\uparrow$ \\
                \midrule
                Keypoint Prediction       & 30.73          & 23.25          & 19.85 \\
                2D Height Uncertainty     & 32.52          & 25.06          & 21.19 \\
                Regress Bias              & 33.84          & 25.73          & 22.30 \\
                \rowcolor{hdColor}
                \textbf{\newDepth (Ours)} & \textbf{35.86} & \textbf{26.22} & \textbf{23.58} \\
                \midrule
            \end{tabular}}
        \label{tab:ablation-more-depth}
    \end{minipage}
    \hfill
    \begin{minipage}{0.48\columnwidth}
        \centering
        \small
        \caption{
            Ablation on VFM selection.
        }
        \setlength{\tabcolsep}{3pt}
        \resizebox{1\linewidth}{!}{
            \begin{tabular}{l|ccc}
                \midrule
                VFM Selection            & Easy$\uparrow$ & Mod.$\uparrow$ & Hard$\uparrow$ \\
                \midrule
                DINOv2                   & 34.73          & 26.05          & 22.55 \\
                SigLIP 2                 & 32.99          & 25.51          & 22.04 \\
                OpenCLIP                 & 35.26          & 25.31          & 22.71 \\
                \rowcolor{hdColor}
                \textbf{EVA-CLIP (Ours)} & \textbf{35.86} & \textbf{26.22} & \textbf{23.58} \\
                \midrule
            \end{tabular}}
        \label{tab:ablation-vfm-selection}
    \end{minipage}

\end{table}

\begin{table}[t]
    \centering

    \begin{minipage}{0.48\columnwidth}
        \centering
        \small
        \caption{
            Ablation on HFA coefficient.
        }
        \setlength{\tabcolsep}{3pt}
        \resizebox{1\linewidth}{!}{
            \begin{tabular}{l|ccc}
                \midrule
                HFA Coefficient & Easy$\uparrow$ & Mod.$\uparrow$ & Hard$\uparrow$ \\
                \midrule
                1               & 33.46          & 25.06          & 21.90 \\
                2               & 34.73          & 25.94          & 22.12 \\
                \rowcolor{hdColor}
                5               & \textbf{35.86} & \textbf{26.22} & \textbf{23.58} \\
                10              & 33.89          & 25.21          & 21.91 \\
                \midrule
            \end{tabular}}
        \label{tab:ablation-alignloss-coe}
    \end{minipage}

\end{table}

\subsection{VFM Selection}
\label{appendix-ablation-vfm-selection}

\cref{tab:ablation-vfm-selection} presents the ablation study on the choice of Vision Foundation Models (VFMs).
We compare several representative pretrained encoders, including DINOv2~\cite{oquab2023dinov2}, SigLIP 2~\cite{tschannen2025siglip}, OpenCLIP~\cite{cherti2023reproducible}, and EVA-CLIP~\cite{sun2023eva}.
Among them, DINOv2 achieves \(34.73\%\), \(26.05\%\), and \(22.55\%\) on Easy, Moderate, and Hard settings, respectively, while SigLIP 2 is lower on all three.
OpenCLIP attains higher Easy and Hard scores than DINOv2 (\(35.26\%\) and \(22.71\%\)) but a lower Moderate score (\(25.31\%\)).
EVA-CLIP delivers the best performance across all difficulty levels, achieving \(35.86\%\), \(26.22\%\), and \(23.58\%\), demonstrating the effectiveness of stronger pretrained visual representations for our framework.
Therefore, EVA-CLIP is adopted as the default VFM in the main experiments. We note that stronger vision--language pretraining does not automatically translate into fine-grained perception, as multimodal LLMs built on such encoders still struggle with fine-grained visual classification~\cite{zhu2026textual}; this motivates distilling VFM features into a task-specific student rather than relying on the VFM alone.

\subsection{Alignment Loss Coefficient}
\label{appendix-ablation-alignment-loss}

\cref{tab:ablation-alignloss-coe} studies the effect of the HFA loss coefficient $\lambda_{align}$.
We vary the coefficient from 1 to 10 and report the detection performance under different difficulty levels.
As the coefficient increases from 1 to 5, the performance consistently improves, reaching the best results at 5 with \(35.86\%\), \(26.22\%\), and \(23.58\%\) on Easy, Moderate, and Hard, respectively.
However, further increasing the coefficient to 10 leads to noticeable performance degradation.
This observation suggests that a moderate coefficient provides a better balance for the distillation, and we therefore set the HFA coefficient to 5 in all experiments.

\section{Limitations}
\label{sec:limitations}
\methodName has three main limitations. First, the closed-form correction in \cref{eq:prodepth} assumes pinhole projection and an upright box, and treats $h_{2D}$ as the tight projection of the eight 3D corners; deviations from this last assumption (truncation, occlusion, annotation tightness) account for the residual error of \newDepth under oracle inputs in \cref{fig:teaser}a. Second, depth accuracy is now coupled to predicted size and yaw, so residual attribute errors, especially on heavily occluded or distant objects, still propagate into depth. Third, our evaluation focuses on outdoor driving benchmarks (\kitti, \nuscenes, \waymo, and \omnidataset), and the behavior of \methodName under substantially different camera setups has not been studied.

\section{Broader Impact}
\label{sec:broader-impact}
\methodName targets \monoThreeD, which underpins applications in autonomous driving, robotics, and augmented reality. By improving depth accuracy without additional sensors or supervision, our work may extend the reach of camera-only 3D perception in settings where LiDAR is impractical or cost-prohibitive. At the same time, any 3D detection technology carries potential for misuse in unauthorized surveillance, and imperfect perception deployed in safety-critical systems can lead to harm. Moreover, because \geoDistill distills representations from a foundation model, the student may inherit spurious correlations or biases present in the teacher~\cite{chen2026safety}. Such inherited behavior is not always visible from accuracy alone, although recent work shows that training-time modifications can leave traces detectable from model outputs~\cite{chen2026unlearning} and that a model's lineage can be traced from its weights~\cite{chen2026who}, which may help audit what a distilled student carries over. We position \methodName as research on geometric depth modeling rather than a deployment-ready system: real-world use should be paired with rigorous validation, redundant sensing where feasible, and adherence to domain-specific safety standards.

\clearpage

\newpage
\section*{NeurIPS Paper Checklist}

\begin{enumerate}

\item {\bf Claims}
    \item[] Question: Do the main claims made in the abstract and introduction accurately reflect the paper's contributions and scope?
    \item[] Answer: \answerYes{}
    \item[] Justification: The abstract and introduction (Sec.~\ref{sec:intro}) precisely state our three contributions: (i) the identification of the front-surface bias in the conventional geometric depth formulation $z = fH/h_{2D}$, (ii) \newDepth, a projection-consistent depth formulation that corrects the bias in closed form, and (iii) \geoDistill, a teacher–student distillation strategy that strengthens attribute perception. Each claim is supported by experimental results in Sec.~\ref{sec:exp}, including the bias diagnosis under oracle inputs (Fig.~\ref{fig:teaser}), \sota results on \kitti, \nuscenes, \waymo, and \omnidataset, and ablation studies on each proposed component. Reported gains and the scope of evaluation faithfully match the claims made.
    \item[] Guidelines:
    \begin{itemize}
        \item The answer \answerNA{} means that the abstract and introduction do not include the claims made in the paper.
        \item The abstract and/or introduction should clearly state the claims made, including the contributions made in the paper and important assumptions and limitations. A \answerNo{} or \answerNA{} answer to this question will not be perceived well by the reviewers.
        \item The claims made should match theoretical and experimental results, and reflect how much the results can be expected to generalize to other settings.
        \item It is fine to include aspirational goals as motivation as long as it is clear that these goals are not attained by the paper.
    \end{itemize}

\item {\bf Limitations}
    \item[] Question: Does the paper discuss the limitations of the work performed by the authors?
    \item[] Answer: \answerYes{}
    \item[] Justification: We discuss three main limitations in Appendix~\ref{sec:limitations}. First, the closed-form correction assumes pinhole projection and an upright box, and treats $h_{2D}$ as the tight projection of the eight \threeD corners. Second, \newDepth couples depth to the predicted \threeD size and yaw, so depth accuracy depends on the accuracy of these attributes; we discuss this dependency in Sec.~\ref{sec:method} and address it through \geoDistill, but residual attribute errors (\eg, on heavily occluded or distant objects with weak visual cues) still propagate to depth. Third, our evaluation focuses on outdoor driving benchmarks (\kitti, \nuscenes, \waymo, and \omnidataset), and the behavior of \methodName under significantly different camera setups has not been studied in this work.
    \item[] Guidelines:
    \begin{itemize}
        \item The answer \answerNA{} means that the paper has no limitation while the answer \answerNo{} means that the paper has limitations, but those are not discussed in the paper.
        \item The authors are encouraged to create a separate ``Limitations'' section in their paper.
        \item The paper should point out any strong assumptions and how robust the results are to violations of these assumptions (e.g., independence assumptions, noiseless settings, model well-specification, asymptotic approximations only holding locally). The authors should reflect on how these assumptions might be violated in practice and what the implications would be.
        \item The authors should reflect on the scope of the claims made, e.g., if the approach was only tested on a few datasets or with a few runs. In general, empirical results often depend on implicit assumptions, which should be articulated.
        \item The authors should reflect on the factors that influence the performance of the approach. For example, a facial recognition algorithm may perform poorly when image resolution is low or images are taken in low lighting. Or a speech-to-text system might not be used reliably to provide closed captions for online lectures because it fails to handle technical jargon.
        \item The authors should discuss the computational efficiency of the proposed algorithms and how they scale with dataset size.
        \item If applicable, the authors should discuss possible limitations of their approach to address problems of privacy and fairness.
        \item While the authors might fear that complete honesty about limitations might be used by reviewers as grounds for rejection, a worse outcome might be that reviewers discover limitations that aren't acknowledged in the paper. The authors should use their best judgment and recognize that individual actions in favor of transparency play an important role in developing norms that preserve the integrity of the community. Reviewers will be specifically instructed to not penalize honesty concerning limitations.
    \end{itemize}

\item {\bf Theory assumptions and proofs}
    \item[] Question: For each theoretical result, does the paper provide the full set of assumptions and a complete (and correct) proof?
    \item[] Answer: \answerNA{}
    \item[] Justification: The paper does not contain formal theoretical results. The geometric derivation of \newDepth in Sec.~\ref{sec:method} relies on standard perspective projection.
    \item[] Guidelines:
    \begin{itemize}
        \item The answer \answerNA{} means that the paper does not include theoretical results.
        \item All the theorems, formulas, and proofs in the paper should be numbered and cross-referenced.
        \item All assumptions should be clearly stated or referenced in the statement of any theorems.
        \item The proofs can either appear in the main paper or the supplemental material, but if they appear in the supplemental material, the authors are encouraged to provide a short proof sketch to provide intuition.
        \item Inversely, any informal proof provided in the core of the paper should be complemented by formal proofs provided in appendix or supplemental material.
        \item Theorems and Lemmas that the proof relies upon should be properly referenced.
    \end{itemize}

    \item {\bf Experimental result reproducibility}
    \item[] Question: Does the paper fully disclose all the information needed to reproduce the main experimental results of the paper to the extent that it affects the main claims and/or conclusions of the paper (regardless of whether the code and data are provided or not)?
    \item[] Answer: \answerYes{}
    \item[] Justification: We provide all the information needed to reproduce our results. The architecture of \methodName, including \newDepth, \mixtureBackbone, and \featureAlignment, is fully described in Sec.~\ref{sec:method} and Appendix~\ref{appendix-imple}, with all training losses and loss weights specified. Implementation details (ResNet-50 backbone, EVA-CLIP as the VFM, \monocop~\cite{zhang2025unleashing} as the base detector, AdamW optimizer with the learning rate schedule, batch size, and hardware setup) are reported in Sec.~\ref{sec:exp} and Appendix~\ref{appendix-imple}. All experiments use publicly available benchmarks (\kitti, \nuscenes, \waymo, and \omnidataset) under their standard splits, with dataset sources cited. The code and trained checkpoints are publicly available at \codeURL.
    \item[] Guidelines:
    \begin{itemize}
        \item The answer \answerNA{} means that the paper does not include experiments.
        \item If the paper includes experiments, a \answerNo{} answer to this question will not be perceived well by the reviewers: Making the paper reproducible is important, regardless of whether the code and data are provided or not.
        \item If the contribution is a dataset and\slash or model, the authors should describe the steps taken to make their results reproducible or verifiable.
        \item Depending on the contribution, reproducibility can be accomplished in various ways. For example, if the contribution is a novel architecture, describing the architecture fully might suffice, or if the contribution is a specific model and empirical evaluation, it may be necessary to either make it possible for others to replicate the model with the same dataset, or provide access to the model. In general. releasing code and data is often one good way to accomplish this, but reproducibility can also be provided via detailed instructions for how to replicate the results, access to a hosted model (e.g., in the case of a large language model), releasing of a model checkpoint, or other means that are appropriate to the research performed.
        \item While NeurIPS does not require releasing code, the conference does require all submissions to provide some reasonable avenue for reproducibility, which may depend on the nature of the contribution. For example
        \begin{enumerate}
            \item If the contribution is primarily a new algorithm, the paper should make it clear how to reproduce that algorithm.
            \item If the contribution is primarily a new model architecture, the paper should describe the architecture clearly and fully.
            \item If the contribution is a new model (e.g., a large language model), then there should either be a way to access this model for reproducing the results or a way to reproduce the model (e.g., with an open-source dataset or instructions for how to construct the dataset).
            \item We recognize that reproducibility may be tricky in some cases, in which case authors are welcome to describe the particular way they provide for reproducibility. In the case of closed-source models, it may be that access to the model is limited in some way (e.g., to registered users), but it should be possible for other researchers to have some path to reproducing or verifying the results.
        \end{enumerate}
    \end{itemize}

\item {\bf Open access to data and code}
    \item[] Question: Does the paper provide open access to the data and code, with sufficient instructions to faithfully reproduce the main experimental results, as described in supplemental material?
    \item[] Answer: \answerYes{}
    \item[] Justification: We release a public repository (\codeURL) containing the full source code of \methodName, configuration files, training scripts, and instructions for environment setup. All datasets used (\kitti, \nuscenes, \waymo, and \omnidataset) are publicly available under their respective licenses, and we follow their standard preprocessing protocols, with details described in Sec.~\ref{sec:exp} and Appendix~\ref{appendix-imple}. The repository includes step-by-step commands to reproduce the main experimental results reported in the paper.
    \item[] Guidelines:
    \begin{itemize}
        \item The answer \answerNA{} means that paper does not include experiments requiring code.
        \item Please see the NeurIPS code and data submission guidelines (\url{https://neurips.cc/public/guides/CodeSubmissionPolicy}) for more details.
        \item While we encourage the release of code and data, we understand that this might not be possible, so \answerNo{} is an acceptable answer. Papers cannot be rejected simply for not including code, unless this is central to the contribution (e.g., for a new open-source benchmark).
        \item The instructions should contain the exact command and environment needed to run to reproduce the results. See the NeurIPS code and data submission guidelines (\url{https://neurips.cc/public/guides/CodeSubmissionPolicy}) for more details.
        \item The authors should provide instructions on data access and preparation, including how to access the raw data, preprocessed data, intermediate data, and generated data, etc.
        \item The authors should provide scripts to reproduce all experimental results for the new proposed method and baselines. If only a subset of experiments are reproducible, they should state which ones are omitted from the script and why.
        \item At submission time, to preserve anonymity, the authors should release anonymized versions (if applicable).
        \item Providing as much information as possible in supplemental material (appended to the paper) is recommended, but including URLs to data and code is permitted.
    \end{itemize}

\item {\bf Experimental setting/details}
    \item[] Question: Does the paper specify all the training and test details (e.g., data splits, hyperparameters, how they were chosen, type of optimizer) necessary to understand the results?
    \item[] Answer: \answerYes{}
    \item[] Justification: All experimental details are reported in Sec.~\ref{sec:exp} and Appendix~\ref{appendix-imple}, including dataset splits for \kitti, \nuscenes, \waymo, and \omnidataset, evaluation metrics and IoU thresholds, the optimizer (AdamW) with its learning rate, weight decay, and decay schedule, the batch size, the number of training epochs, the loss weights for all training objectives, and the hardware setup. Hyperparameters follow the conventions of the base detector~\cite{zhang2025unleashing} and are not separately tuned for our method.
    \begin{itemize}
        \item The answer \answerNA{} means that the paper does not include experiments.
        \item The experimental setting should be presented in the core of the paper to a level of detail that is necessary to appreciate the results and make sense of them.
        \item The full details can be provided either with the code, in appendix, or as supplemental material.
    \end{itemize}

\item {\bf Experiment statistical significance}
    \item[] Question: Does the paper report error bars suitably and correctly defined or other appropriate information about the statistical significance of the experiments?
    \item[] Answer: \answerNo{}
    \item[] Justification: We do not report error bars. Following the standard evaluation protocol of \monoThreeD on \kitti, \nuscenes, \waymo, and \omnidataset~\cite{zhang2023monodetr, pu2024monodgp, zhang2025unleashing, brazil2023omni3d}, results are obtained from a single training run per configuration, since each run requires substantial compute (eight NVIDIA A100 GPUs). Repeated runs across all benchmarks and ablation settings would be computationally prohibitive. To compensate, we evaluate \methodName on four diverse benchmarks and observe consistent improvements over strong baselines across datasets, difficulty levels, and detector architectures (\cref{fig:prodepth-generalization}), which we believe supports the statistical reliability of our reported gains.
    \item[] Guidelines:
    \begin{itemize}
        \item The answer \answerNA{} means that the paper does not include experiments.
        \item The authors should answer \answerYes{} if the results are accompanied by error bars, confidence intervals, or statistical significance tests, at least for the experiments that support the main claims of the paper.
        \item The factors of variability that the error bars are capturing should be clearly stated (for example, train/test split, initialization, random drawing of some parameter, or overall run with given experimental conditions).
        \item The method for calculating the error bars should be explained (closed form formula, call to a library function, bootstrap, etc.)
        \item The assumptions made should be given (e.g., Normally distributed errors).
        \item It should be clear whether the error bar is the standard deviation or the standard error of the mean.
        \item It is OK to report 1-sigma error bars, but one should state it. The authors should preferably report a 2-sigma error bar than state that they have a 96\% CI, if the hypothesis of Normality of errors is not verified.
        \item For asymmetric distributions, the authors should be careful not to show in tables or figures symmetric error bars that would yield results that are out of range (e.g., negative error rates).
        \item If error bars are reported in tables or plots, the authors should explain in the text how they were calculated and reference the corresponding figures or tables in the text.
    \end{itemize}

\item {\bf Experiments compute resources}
    \item[] Question: For each experiment, does the paper provide sufficient information on the computer resources (type of compute workers, memory, time of execution) needed to reproduce the experiments?
    \item[] Answer: \answerYes{}
    \item[] Justification: All experiments are conducted on a single node with eight NVIDIA A100 GPUs (80GB) and a batch size of 16, as reported in Appendix~\ref{appendix-imple}. A typical full training run on \kitti takes approximately 12 hours, while training on the larger \waymo and \omnidataset benchmarks takes one to two days. Inference cost is reported in \cref{tab:eff}, where \methodName runs at 62.5 GFLOPs with 42.0M parameters, slightly more efficient than the \monocop~\cite{zhang2025unleashing} baseline. We acknowledge that the full research project, including preliminary explorations and failed designs, required compute beyond the final reported experiments.
    \item[] Guidelines:
    \begin{itemize}
        \item The answer \answerNA{} means that the paper does not include experiments.
        \item The paper should indicate the type of compute workers CPU or GPU, internal cluster, or cloud provider, including relevant memory and storage.
        \item The paper should provide the amount of compute required for each of the individual experimental runs as well as estimate the total compute.
        \item The paper should disclose whether the full research project required more compute than the experiments reported in the paper (e.g., preliminary or failed experiments that didn't make it into the paper).
    \end{itemize}

\item {\bf Code of ethics}
    \item[] Question: Does the research conducted in the paper conform, in every respect, with the NeurIPS Code of Ethics \url{https://neurips.cc/public/EthicsGuidelines}?
    \item[] Answer: \answerYes{}
    \item[] Justification: We have reviewed the NeurIPS Code of Ethics, and our research fully conforms to its guidelines. The work uses publicly available autonomous driving benchmarks (\kitti, \nuscenes, \waymo, and \omnidataset) under their respective licenses, does not involve human subjects or private data.
    \item[] Guidelines:
    \begin{itemize}
        \item The answer \answerNA{} means that the authors have not reviewed the NeurIPS Code of Ethics.
        \item If the authors answer \answerNo, they should explain the special circumstances that require a deviation from the Code of Ethics.
        \item The authors should make sure to preserve anonymity (e.g., if there is a special consideration due to laws or regulations in their jurisdiction).
    \end{itemize}

\item {\bf Broader impacts}
    \item[] Question: Does the paper discuss both potential positive societal impacts and negative societal impacts of the work performed?
    \item[] Answer: \answerYes{}
    \item[] Justification: \methodName improves monocular \threeD object detection, which has direct positive applications in autonomous driving, robotics, and augmented reality, where reliable depth perception from a single camera reduces dependency on expensive LiDAR sensors and broadens accessibility to safety-critical perception. As with any \threeD detection technology, potential negative impacts include misuse in unauthorized surveillance and the risk of perception errors in deployed autonomous systems leading to safety hazards. We emphasize that \methodName is intended as research progress on geometric depth modeling rather than a deployment-ready system, and any real-world use should be accompanied by rigorous validation, redundant sensing modalities, and adherence to relevant safety standards.
    \item[] Guidelines:
    \begin{itemize}
        \item The answer \answerNA{} means that there is no societal impact of the work performed.
        \item If the authors answer \answerNA{} or \answerNo, they should explain why their work has no societal impact or why the paper does not address societal impact.
        \item Examples of negative societal impacts include potential malicious or unintended uses (e.g., disinformation, generating fake profiles, surveillance), fairness considerations (e.g., deployment of technologies that could make decisions that unfairly impact specific groups), privacy considerations, and security considerations.
        \item The conference expects that many papers will be foundational research and not tied to particular applications, let alone deployments. However, if there is a direct path to any negative applications, the authors should point it out. For example, it is legitimate to point out that an improvement in the quality of generative models could be used to generate Deepfakes for disinformation. On the other hand, it is not needed to point out that a generic algorithm for optimizing neural networks could enable people to train models that generate Deepfakes faster.
        \item The authors should consider possible harms that could arise when the technology is being used as intended and functioning correctly, harms that could arise when the technology is being used as intended but gives incorrect results, and harms following from (intentional or unintentional) misuse of the technology.
        \item If there are negative societal impacts, the authors could also discuss possible mitigation strategies (e.g., gated release of models, providing defenses in addition to attacks, mechanisms for monitoring misuse, mechanisms to monitor how a system learns from feedback over time, improving the efficiency and accessibility of ML).
    \end{itemize}

\item {\bf Safeguards}
    \item[] Question: Does the paper describe safeguards that have been put in place for responsible release of data or models that have a high risk for misuse (e.g., pre-trained language models, image generators, or scraped datasets)?
    \item[] Answer: \answerNA{}
    \item[] Justification: Our work releases a monocular \threeD object detector for outdoor driving scenes. The model is not a generative system, does not produce or scrape data, and does not pose a high risk of misuse such as disinformation or harmful content generation. The released code and checkpoints are intended for research purposes on standard public benchmarks, and therefore no additional safeguards beyond those discussed in the broader-impact statement are required.
    \item[] Guidelines:
    \begin{itemize}
        \item The answer \answerNA{} means that the paper poses no such risks.
        \item Released models that have a high risk for misuse or dual-use should be released with necessary safeguards to allow for controlled use of the model, for example by requiring that users adhere to usage guidelines or restrictions to access the model or implementing safety filters.
        \item Datasets that have been scraped from the Internet could pose safety risks. The authors should describe how they avoided releasing unsafe images.
        \item We recognize that providing effective safeguards is challenging, and many papers do not require this, but we encourage authors to take this into account and make a best faith effort.
    \end{itemize}

\item {\bf Licenses for existing assets}
    \item[] Question: Are the creators or original owners of assets (e.g., code, data, models), used in the paper, properly credited and are the license and terms of use explicitly mentioned and properly respected?
    \item[] Answer: \answerYes{}
    \item[] Justification: All existing assets used in this work are properly cited and used in accordance with their licenses. We use four public benchmarks: \kitti~\cite{geiger2012we} (CC BY-NC-SA 3.0), \nuscenes~\cite{caesar2020nuscenes} (CC BY-NC-SA 4.0), \waymo Open Dataset~\cite{sun2020scalability} (Waymo Dataset License Agreement for non-commercial use), and \omnidataset~\cite{brazil2023omni3d} (CC BY-NC 4.0), all employed under their non-commercial research terms. Our implementation builds on \monocop~\cite{zhang2025unleashing} and adopts CLIP~\cite{sun2023eva} as the Vision Foundation Model; both are released for research use and properly credited in Sec.~\ref{sec:exp} and Appendix~\ref{appendix-imple}.
    \item[] Guidelines:
    \begin{itemize}
        \item The answer \answerNA{} means that the paper does not use existing assets.
        \item The authors should cite the original paper that produced the code package or dataset.
        \item The authors should state which version of the asset is used and, if possible, include a URL.
        \item The name of the license (e.g., CC-BY 4.0) should be included for each asset.
        \item For scraped data from a particular source (e.g., website), the copyright and terms of service of that source should be provided.
        \item If assets are released, the license, copyright information, and terms of use in the package should be provided. For popular datasets, \url{paperswithcode.com/datasets} has curated licenses for some datasets. Their licensing guide can help determine the license of a dataset.
        \item For existing datasets that are re-packaged, both the original license and the license of the derived asset (if it has changed) should be provided.
        \item If this information is not available online, the authors are encouraged to reach out to the asset's creators.
    \end{itemize}

\item {\bf New assets}
    \item[] Question: Are new assets introduced in the paper well documented and is the documentation provided alongside the assets?
    \item[] Answer: \answerYes{}
    \item[] Justification: The new asset introduced in this work is the source code of \methodName, including the implementation of \newDepth, \mixtureBackbone, and \featureAlignment. We release a public repository (\codeURL) containing the source code, configuration files, training and evaluation scripts, environment setup instructions, and a README documenting usage, together with the trained checkpoints.
    \item[] Guidelines:
    \begin{itemize}
        \item The answer \answerNA{} means that the paper does not release new assets.
        \item Researchers should communicate the details of the dataset\slash code\slash model as part of their submissions via structured templates. This includes details about training, license, limitations, etc.
        \item The paper should discuss whether and how consent was obtained from people whose asset is used.
        \item At submission time, remember to anonymize your assets (if applicable). You can either create an anonymized URL or include an anonymized zip file.
    \end{itemize}

\item {\bf Crowdsourcing and research with human subjects}
    \item[] Question: For crowdsourcing experiments and research with human subjects, does the paper include the full text of instructions given to participants and screenshots, if applicable, as well as details about compensation (if any)?
    \item[] Answer: \answerNA{}
    \item[] Justification: This work does not involve crowdsourcing or research with human subjects. All experiments use existing public benchmarks for monocular \threeD object detection.
    \item[] Guidelines:
    \begin{itemize}
        \item The answer \answerNA{} means that the paper does not involve crowdsourcing nor research with human subjects.
        \item Including this information in the supplemental material is fine, but if the main contribution of the paper involves human subjects, then as much detail as possible should be included in the main paper.
        \item According to the NeurIPS Code of Ethics, workers involved in data collection, curation, or other labor should be paid at least the minimum wage in the country of the data collector.
    \end{itemize}

\item {\bf Institutional review board (IRB) approvals or equivalent for research with human subjects}
    \item[] Question: Does the paper describe potential risks incurred by study participants, whether such risks were disclosed to the subjects, and whether Institutional Review Board (IRB) approvals (or an equivalent approval/review based on the requirements of your country or institution) were obtained?
    \item[] Answer: \answerNA{}
    \item[] Justification: This work does not involve human subjects, so IRB approval is not applicable.
    \item[] Guidelines:
    \begin{itemize}
        \item The answer \answerNA{} means that the paper does not involve crowdsourcing nor research with human subjects.
        \item Depending on the country in which research is conducted, IRB approval (or equivalent) may be required for any human subjects research. If you obtained IRB approval, you should clearly state this in the paper.
        \item We recognize that the procedures for this may vary significantly between institutions and locations, and we expect authors to adhere to the NeurIPS Code of Ethics and the guidelines for their institution.
        \item For initial submissions, do not include any information that would break anonymity (if applicable), such as the institution conducting the review.
    \end{itemize}

\item {\bf Declaration of LLM usage}
    \item[] Question: Does the paper describe the usage of LLMs if it is an important, original, or non-standard component of the core methods in this research? Note that if the LLM is used only for writing, editing, or formatting purposes and does \emph{not} impact the core methodology, scientific rigor, or originality of the research, declaration is not required.
    \item[] Answer: \answerNA{}
    \item[] Justification: LLMs are not used as part of the core methodology of this research. The proposed \methodName, including \newDepth and \geoDistill, does not involve any LLM component.
    \item[] Guidelines:
    \begin{itemize}
        \item The answer \answerNA{} means that the core method development in this research does not involve LLMs as any important, original, or non-standard components.
        \item Please refer to our LLM policy in the NeurIPS handbook for what should or should not be described.
    \end{itemize}

\end{enumerate}

\end{document}